\documentclass[journal,trans]{IEEEtran}

\ifCLASSINFOpdf
\else
\fi

\usepackage{epsfig} 
\usepackage{cite} 
\usepackage{amsmath,amssymb} 
\usepackage{algorithm,algpseudocode,algorithmicx}
\usepackage[mathscr]{eucal}
\usepackage{amsfonts,amssymb}
\usepackage{graphicx,subfigure,cases,multirow,graphics}
\usepackage[colorlinks,citecolor=blue,urlcolor=black,linkcolor=blue]{hyperref}

\newlength\savewidth
\newcommand\shline{\noalign{\global\savewidth\arrayrulewidth\global\arrayrulewidth 1.0pt}\hline\noalign{\global\arrayrulewidth\savewidth}}

\newlength\savedwidth

\newcommand{\bS}{\mathbb S}
\newcommand{\bR}{\mathbb R}
\newcommand{\bZ}{\mathbb Z}
\newcommand{\bM}{\mathbb M}

\newcommand{\cJ}{\mathcal J}
\newcommand{\cC}{\mathcal C}

\newcommand{\cL}{\mathcal L}
\newcommand{\cF}{\mathcal F}
\newcommand{\cT}{\mathcal T}
\newcommand{\cD}{\mathcal D}
\newcommand{\cU}{\mathcal U}
\newcommand{\cG}{\mathcal G}

\newcommand{\cE}{\mathcal E}
\newcommand{\kD}{\mathfrak D}
\newcommand{\kC}{\mathfrak C}
\newcommand{\kF}{\mathfrak F}
\newcommand{\rC}{\mathscr C}
\newcommand{\rT}{\mathscr T}
\DeclareMathOperator\Lip{Lip}

\newcommand{\ta}{\tilde a}
\newcommand{\tx}{\tilde x}

\newcommand{\tp}{\tilde p}
\newcommand{\fu}{\mathbf u}
\newcommand{\fv}{\mathbf v}
\newcommand{\fn}{\mathbf n}

\newcommand{\algorithmiclastcon}{\textbf{Initialization:}}
\newcommand{\lastcon}{\item[\algorithmiclastcon]}

\newcommand{\algorithmicdotdef}{\textbf{$\bullet$}}
\newcommand{\dotdef}{\item[\algorithmicdotdef]}

\begin{document}
%
\title{Trajectory Grouping with Curvature Regularization for Tubular Structure Tracking}


\author{\IEEEauthorblockN{Li Liu, Da Chen, Minglei Shu, Baosheng Li, Huazhong Shu, Michel Paques and Laurent D. Cohen,~\IEEEmembership{Fellow,~IEEE}}
\IEEEcompsocitemizethanks{
\IEEEcompsocthanksitem Li~Liu, Da~Chen and Minglei Shu are with Qilu University of Technology (Shandong Academy of Sciences), Shandong Artificial Intelligence Institute, 250014 Jinan, China. (email: dachen.cn@hotmail.com)
\IEEEcompsocthanksitem Baosheng Li is with Department of Radiation Oncology (Chest Section), Shandong's Key Laboratory of Radiation Oncology, Shandong Cancer Hospital, Shandong Academy of Medical Sciences, Jinan, China; Department of Radiation Oncology, Shandong Cancer Hospital \& Institute, Shandong Academy of Medical Sciences, Jinan, China
\IEEEcompsocthanksitem Huazhong Shu is Jiangsu Provincial Joint International Research Laboratory of Medical Information Processing, School of Computer Science and Engineering, Southeast University, Nanjing, 210096, China.
\IEEEcompsocthanksitem Michel~Paques is with the Centre Hospitalier National dOphtalmologie des Quinze-Vingts, Paris, France.
\IEEEcompsocthanksitem Laurent~D.~Cohen is with University Paris Dauphine, PSL Research University, CNRS, UMR 7534, CEREMADE, 75016 Paris, France.}
\thanks{Li~Liu and Da~Chen are the co-first authors with equal contributions. Minglei~Shu~(email:~shuml@sdas.org) is the corresponding author.}}

\maketitle

\markboth{Journal of \LaTeX\ Class Files,~Vol.~13, No.~9, September~2020}%
{Shell \MakeLowercase{\textit{et al.}}: Bare Demo of IEEEtran.cls for Journals}
%



\IEEEtitleabstractindextext{%
\begin{abstract}
Tubular structure tracking is a crucial task in the fields of computer vision and medical image analysis. The minimal paths-based approaches have exhibited their strong ability in tracing tubular structures, by which a tubular structure can be naturally modeled as a minimal geodesic path computed with a suitable geodesic metric. However, existing minimal paths-based tracing approaches still suffer from difficulties such as the shortcuts and short branches combination problems, especially when dealing with the images involving complicated tubular tree structures or background. In this paper, we introduce a new minimal paths-based model for minimally interactive tubular structure centerline extraction in conjunction with a perceptual grouping scheme. Basically, we take into account the prescribed tubular trajectories and curvature-penalized geodesic paths to seek suitable shortest paths. The proposed approach can benefit from the local smoothness prior on tubular structures and the global optimality of the used graph-based path searching scheme. Experimental results on both synthetic and real images prove that the proposed model indeed obtains outperformance comparing with the state-of-the-art minimal paths-based tubular structure tracing algorithms.
\end{abstract}

\begin{IEEEkeywords}
Tubular structure tracking, minimal path, perceptual grouping, curvature penalization, fast marching algorithm, graph optimization.
\end{IEEEkeywords}}

\IEEEdisplaynontitleabstractindextext

%
\IEEEpeerreviewmaketitle

\section{Introduction}
Tracing tubular structures such as blood vessels, roads and rivers is a fundamental task  arisen in the fields of computer vision,  medical imaging and remote sensing. A basic objective for tubular structure  tracking is to search for the centerline and/or the tubular boundaries in both sides to delineate an elongated structure. This is very often carried out by investigating  the tubular anisotropy and appearance features to identify the centerline positions. These tubular features in general can be extracted through various multi-scale and multi-orientation filters as reviewed in~\cite{moccia2018blood,lesage2009review}.
The existing tubular structure tracking approaches can be roughly divided into two categories: (i) automatic tracking models for which all the branches are expected to be identified, and (ii) interactive tracking models where the user-intervention is often taken into consideration. In this paper, we focus on the minimally interactive tubular structure tracking approaches.

\begin{figure*}[t]
\centering
\includegraphics[width=17cm]{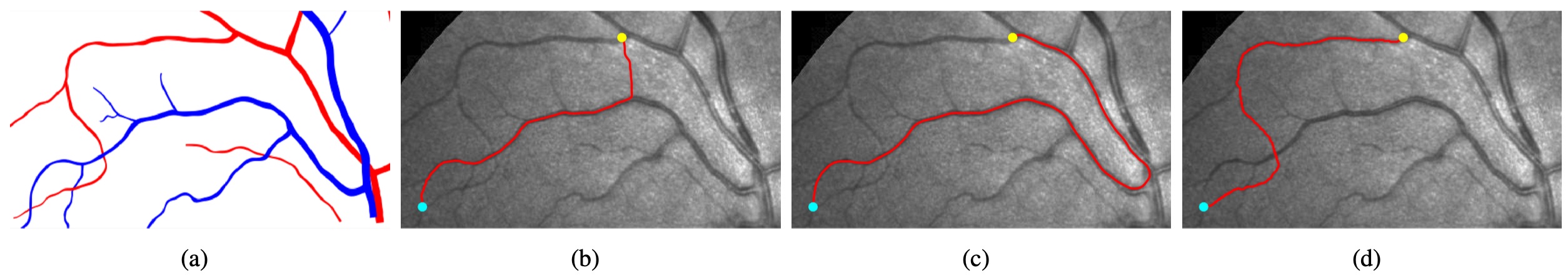}
\caption{Examples for shortcuts and short branches combination problems. \textbf{a} The regions indicated by red and blue color represent the retinal artery and vein vessels.  The objective is to extract the artery vessel between two given points indicated by dots. \textbf{b} to \textbf{d} Vessel extraction results via anisotropic minimal path model~\cite{benmansour2011tubular}, the Finsler variant of the Sub-Riemannian minimal path model~\cite{duits2018optimal,mirebeau2018fast} and the proposed method, respectively. The shortcuts and short branches combination problems are observed in figures \textbf{b} and \textbf{c}, while the geodesic path derived from the proposed model indeed seeks the correct vessel}
\label{fig:examples}	
\end{figure*}

A simple and effective idea for automatic  tubular structure tracking is implemented by a path growing method. The centerline of each vessel branch is depicted by a locally optimal path propagated from a set of seed points in conjunction with a local tubular features detection procedure~\cite{bekkers2014multi,cetin2012vessel,cetin2015higher}.  Unfortunately, the path growing approaches may fail to detect tubular structures in the presence of gaps, since the objective path can only advance a small step. 
The implementation of minimal paths is an alternative solution for tracking a connected tubular structure tree. Significant examples include the keypoints-based minimal path growing models~\cite{li20093d,bekkers2018nilpotent},  where new source points are iteratively added during the geodesic distance computation. The geodesic voting methods~\cite{rouchdy2013geodesic,chen2016curve} for which the tubular tree can be identified via voting scores, and the minimum spanning tree model~\cite{moriconi2017vtrails} where a tubular structure tree can be identified by finds saddle points from the geodesic distance maps~\cite{cohen2001grouping}.
Other interesting tubular structure centerline tracing approaches include the curve evolution-based models~\cite{mohan2010tubular,wang2011novel}, the tracing algorithms relying on prescribed trajectories~\cite{vandini2017robust,xu2011vessel,lowell2004measurement} and the learning-based tubularity tracking models~\cite{turetken2016reconstructing,de2015graph}. 

Even through they have been extensively studied, the semi- or fully automatic tubularity tracking models still  lack sufficient accuracy and reliability, especially in the case of complex scenario. As an alternative solution,  the type of interactive tubular centerline tracing approaches very often relies on the user intervention such as seed points which define the source and end points for tubular branch. The minimal path models, first introduced by~\cite{cohen1997global}, are regarded as  one of the most successful tools in tracing tubular structures. However, in its original formulation~\cite{cohen1997global}, there is no guarantee that the minimal paths pass through the exact tubular centerlines. In order to address this issue, a two-stage procedure~\cite{deschamps2001fast} is proposed to get the exact centerlines by taking into account the tubularity segmentation to generate centralized potential. A significant improvement on tracing the centerlines and boundaries has been made by~\cite{li2007vessels,benmansour2011tubular}, where an abstract dimension representing the thickness of tubular structures is added, thus a 2D (resp. 3D) vessel can be described by a 3D (resp. 4D) minimal path. However, the short branches combination and shortcuts problems may often occur for these minimal path models, due to the complicated situation. The minimal path models~\cite{chen2019minimal,liao2018progressive} with a dynamic metric update scheme incorporate the update procedure of geodesic metrics during the fronts propagation. The curvature regularization is introduced to minimal path computation in both continuous domain~\cite{chen2017global,duits2018optimal} and discrete domain~\cite{ulen2015shortest}, leading to geodesic paths with rigidity prior to reduce the risk of short branches combination and shortcuts problems. Unfortunately, the existing minimal path models based on the Eikonal partial differential equation (PDE) framework mentioned above are difficult to benefit from the prescribed trajectories and may cost expensive computation burden in the sense of interactive tubular structure tracking. Despite the efforts on the improvement of minimal path techniques, the short branches combination problem still occurs when dealing with complicated situation, as depicted in Fig.~\ref{fig:examples}. Figs.~\ref{fig:examples}b and~\ref{fig:examples}c present the results derived from the anisotropic model~\cite{benmansour2011tubular} and the progressive model~\cite{liao2018progressive}, where one can observe short branches combination issues. While the proposed model indeed obtains good results, see Fig.~\ref{fig:examples}d.
It is worth to point out the graph-based shortest path methods~\cite{poon2007live,wang2013interactive} for tubular trajectory tracing also obtain promising results.  
  
In this paper, we propose a new approach based on minimal paths for minimally interactive tubular structure centerline tracing. It is able to blend the benefits from both of the curvature-penalized geodesic paths and the prescribed tubular trajectories. These trajectories are taken as candidate segments to make up of the tubular centerlines and can be derived from any tubular structure segmentation algorithm. The target shortest paths used to delineate tubular structures are obtained by the Dijkstra's algorithm~\cite{dijkstra1959note} which is established over a graph consisting of nodes and edges. We propose a new and reasonable way to build the weight for each edge in conjunction with curvature-penalized geodesic distance, which forms the main contribution of this paper. 
In~\cite{wang2013interactive}, the authors present a shortest path-based tubular structure tracking model, which also relies on the prescribed trajectories. However, the proposed method differs from the one introduced in~\cite{wang2013interactive} mainly in the way of establishing the connection between two trajectories which likely belong to the same tubular structure.  Specifically, the model in~\cite{wang2013interactive} connects two neighbouring trajectories using a straight segment and measures the connection cost by the Euclidean length of the segment and the related angles between them. However, such a connection scheme may accumulate the approximation errors of bridging the gaps between adjacent trajectories, especially when extracting long structures. In order to address this problem, we consider to complete the gap between two neighbouring trajectories using  a curvature-penalized geodesic path, which is more accurate and natural than using straight segments~\cite{wang2013interactive}. 

The manuscript is organized as follows. In Section~\ref{sec:MPs}, we briefly introduce the background on the computation of curvature-penalized geodesic distances and the corresponding minimal paths. Section~\ref{sec:PG} presents a new tubular structure tracking model, based on the curvature-penalized minimal paths and perceptual grouping. The experimental results and the conclusion presented in Sections~\ref{sec:Exp} and \ref{sec:conclusion}, respectively.

\section{Background on Curvature-Penalized Minimal Path Models}
\label{sec:MPs}
The original isotropic minimal path model~\cite{cohen1997global} is designed to search for the global minimum of a weighted curve length. The curvature-penalized minimal path approaches, such as the Finsler elastica (FE) model~\cite{chen2017global,chen2016new} and the Finsler variant of the sub-Riemannian (FSR) model~\cite{duits2018optimal}, are regarded as two elegant extensions to the original model~\cite{cohen1997global}. In both approaches, the use of the curvature regularization is able to yield geodesic paths with strongly smooth and rigid appearance. 

\subsection{Curvature-Penalized Minimal Paths}
Let $\Omega\subset\bR^2$ be an open and bounded image domain and let $\Lip([0,1],\Omega)$ denote the set of  curves $\gamma:[0,1]\to\Omega$ with Lipschitz continuity. An energy functional, or the weighted curve length,  of  a  curve $\gamma\in\Lip([0,1],\Omega)$ encoding curvature penalization can be formulated as
\begin{equation}
\label{eq:CurvatureLength}
\cL_0(\gamma)=\int_0^1\kC_0(\gamma(u),\gamma^\prime(u))\psi(\kappa(u))\|\gamma^\prime(u)\|du,
\end{equation}
where $\kappa:[0,1]\to\bR$ stands for the curvature of $\gamma$, and $\gamma^\prime=d\gamma/du$ is the first-order derivative of $\gamma$. One can see that the energy~\eqref{eq:CurvatureLength} involves two cost functions $\kC_0$ and $\psi$. Specifically, the  function $\kC_0:\Omega\times\bR^2\to\bR^+$ can be derived from the image data using a steerable filter. Basically, $\kC_0(x,\mathbf{v})$ is expected to have low values if the point $x$ is close to a tubular centerline and the direction $\mathbf{v}$ is proportional to the orientation that a tubular structure should have at $x$. Furthermore, for the FE  and  FSR minimal path models, the cost function $\psi$ can be respectively formulated as $\psi:=\psi_{\rm FE}$ and $\psi:=\psi_{\rm FSR}$ such that
\begin{equation*}
\psi_{\rm FE}(\kappa)=1+(\beta\kappa)^2,~\text{and~} \psi_{\rm FSR}(\kappa)=\sqrt{1+(\beta\kappa)^2},
\end{equation*}
where $\beta\in\bR^+$ is a scalar parameter that controls the importance of the curvature $\kappa$. 

Let $\tilde\Omega=\Omega\times\bS^1$ be an orientation-lifted space of higher dimension, where $\bS^1=[0,2\pi)$ is an interval with periodic boundary condition, and denote by $\fn_\theta=(\cos\theta,\sin\theta)^\top$ a unit vector of an angle $\theta\in\bS^1$. As discussed in~\cite{chen2017global,duits2018optimal}, a key idea for minimizing the energy~\eqref{eq:CurvatureLength} is to lift a planar curve $\gamma\in\Lip([0,1],\Omega)$ to the space $\tilde\Omega$ in conjunction with a parametric function $\tau:[0,1]\to\bS^1$ defined being such that
\begin{equation}
\label{eq_TurningAnkle}
\gamma^\prime=\fn_{\tau}\,\|\gamma^\prime\|,
\end{equation}
yielding that
\begin{equation}
\label{eq:kappa}
\kappa=\frac{\tau^\prime}{\|\gamma^\prime\|}.
\end{equation}
The orientation lifting of $\gamma$ yields a new curve $\tilde\gamma=(\gamma,\tau)$ subject to Eq.~\eqref{eq_TurningAnkle} and its first-oder derivative obeys $\tilde\gamma^\prime(u)=(\gamma^\prime(u),\tau^\prime(u))$ for any $u\in[0,1]$.
Note that each point $(x,\theta)\in\tilde\Omega$ lying in an orientation-lifted curve $\tilde\gamma=(\gamma,\tau)$ has three coordinates, where the first two coordinate $x\in\Omega$ indicates the physical positions and the third coordinate $\theta\in\bS^1$ characterizes the tangent vector $\gamma^\prime(u)$.
With these definitions in hands, the minimization of the energy~\eqref{eq:CurvatureLength} can be efficiently addressed by seeking the minimal curve length of orientation-lifted curves measured through an orientation-lifted Finsler metric $\cF_{\epsilon}:\tilde\Omega\times\bR^3\to\bR^+$, which implicitly encodes the curvature terms. Specifically, for any point $\tx=(x,\theta)\in\tilde\Omega$ and any vector $\tilde\fu=(\fu,\nu)\in\bR^3$, a general form of $\cF_{\epsilon}$ can be expressed as
\begin{equation}
\label{eq:CurvaMetrics}
\cF_{\epsilon}(\tx,\tilde\fu)=\kC(\tx)\kF_{\epsilon}(\tx,\tilde\fu),
\end{equation}
where $\kC:\tilde\Omega\to\bR^+$ is an orientation-dependent cost function derived from image data, subject to $\kC(x,\theta)=\kC_0(x,\fn_\theta)$. The computation for the cost function $\kC$ is presented in Section~\ref{subsec_CostFunction}. In addition, the metric $\kF_{\epsilon}$ only  involves the curvature penalty, thus independent to image data. 

Specifically,  the metric $\kF_{\epsilon}$ for the FE minimal path model~\cite{chen2017global} reads
\begin{equation}
\label{eq:FE}
\kF_{\epsilon}(\tx,\tilde\fu)=\sqrt{\epsilon^{-2}\|\fu\|^2+2\epsilon^{-1}|\beta\nu|^2}+(\epsilon^{-1}-1)\langle \fu,\fn_\theta\rangle.
\end{equation}
While for the FSR minimal path model~\cite{duits2018optimal}, one has
\begin{align}
\label{eq:FSR}
\kF_{\epsilon}(\tx,\tilde\fu)^2=&|\langle\fu,\fn_\theta \rangle|^2+|\beta\nu|^2+\epsilon^{-2}(\|\fu\|^2-|\langle \fu,\fn_\theta\rangle|^2)\nonumber\\
&+(\epsilon^{-2}-1)(\min\{0,\langle\fu,\fn_\theta\rangle\})^2.
\end{align}

For an orientation-lifted curve $\tilde\gamma$, the energy of  $\cL_\epsilon(\tilde\gamma)$ measured by the metric $\cF_\epsilon$ can be expressed as follows:
\begin{equation} 
\label{eq_CurveLength}
\cL_\epsilon(\tilde\gamma)=\int_0^1\cF_\epsilon(\tilde\gamma(u),\tilde\gamma^\prime(u))du.	
\end{equation}
As discussed in~\cite{chen2017global,duits2018optimal}, the minimum of the energy $\cL_0(\gamma)$ in Eq.~\eqref{eq:CurvatureLength} can be well approximated by the length of a geodesic path associated to the metric $\cF_{\epsilon}$ as $\epsilon\to0$, providing that $\kC_0(x,\fn_\theta)=\kC(x,\theta)$. In this way, the minimization problem of the energy $\cL_0(\gamma)$ is transferred to minimizing the new energy $\cL_\epsilon(\tilde\gamma)$, where the later problem can be efficiently addressed by the eikonal PDE framework. 

The geodesic distance map very often lends itself for the minimization of the length $\cL_\epsilon$. For a fixed source point $\ta\in\tilde\Omega$, the geodesic distance map defines a minimal curve length for each point $\tx\in\tilde\Omega$ 
\begin{equation}
\label{eq_dist}
\cU_{\ta}(\tx)=\inf\{\cL_\epsilon(\tilde\gamma)\mid\tilde\gamma\in\Lip([0,1],\tilde\Omega),\,\tilde\gamma(0)=\ta,\tilde\gamma(1)=\tx\}.
\end{equation}
It is well known that the geodesic distance map $\cU_{\ta}$ satisfies the Eikonal equation such that $\cU_{\ta}(\ta)=0$ and for any orientation-lifted point $\tx\in\tilde\Omega\backslash\{\ta\}$ we have
\begin{equation}
\label{eq:EikonalPDE}
\max_{\tilde\fv\neq\mathbf{0}}\frac{\langle\nabla\cU_{\ta}(\tx),\tilde\fv\rangle}{\cF_{\epsilon}(\tx,\tilde\fv)}=1.
\end{equation}
The eikonal equation~\eqref{eq:EikonalPDE} can be solved by using the state-of-the-art Hamiltonian fast marching method~\cite{mirebeau2018fast}, in terms of a Hamiltonian reformulation of Eq.~\eqref{eq:EikonalPDE} .
A geodesic path $\cC_{\ta,\tx}$ linking from $\ta$ to $\tx$ can be derived by re-parametering the solution $\cC$ (which is also a geodesic path) to a gradient descent ordinary differential equation (ODE) on the geodesic distance map $\cU_{\ta}$
\begin{equation}
\label{eq:ODE}
\cC^\prime(u)=-\underset{\|\tilde\fv\|=1}{\arg\max}\left\{ \frac{\langle\tilde\fv,\nabla\cU_{\ta}(\cC(u)) \rangle}{\cF_{\epsilon}(\tilde\fv,\nabla\cU_{\ta}(\cC(u)))}\right\}.
\end{equation}

For two given points with tangents, the minimal paths associated to the data-driven curvature-penalized metric $\cF_{\epsilon}$ attempt to keep smooth, since $\cF_{\epsilon}$ implicitly encodes curvature penalization as regularization. In next section, we present the construction details for the curvature-penalized metric $\cF_{\epsilon}$.

\subsection{Computing the data-driven cost function $\kC$}
\label{subsec_CostFunction}
The function $\kC$ can be estimated from the image data  via a steerable filter~\cite{chen2017global}. In the context of tubular structure tracking, the optimally oriented flux (OOF) filter~\cite{law2008three} is an effective tool for extracting geometry features from images, which will be adopted in this paper. For clarity, we assume that the tubular structures are supported to have \emph{locally lower intensities} than background. 

Let $G_\sigma$ be a Gaussian kernel with standard deviation $\sigma$ and let $\{\partial_{x_ix_j}G_\sigma\}_{i,j}$ be the Hessian matrix of the kernel $G_\sigma$. The response of the OOF filter on a gray level image $I:\Omega\to\bR$ at a point $x$ and a scale $r\in[R_{\rm min},R_{\rm max}]$ is a symmetric matrix of size $2\times2$
\begin{equation}
\label{eq:OS}
\Psi(x,r)=\left(I\ast
\begin{pmatrix}
\partial_{x_1x_1}G_\sigma,&\partial_{x_1x_2}G_\sigma\\	
\partial_{x_2x_1}G_\sigma,&\partial_{x_2x_2}G_\sigma
\end{pmatrix}\ast\chi_r\right)(x),
\end{equation}
where  ``$\ast$'' stands for the convolution operator. The function $\chi_r:\Omega\to\{0,1\}$ is the indicator for a disk of radius  $r$ 
\begin{equation}
\chi_r(x)=
\begin{cases}
1,& \Arrowvert x \Arrowvert\leq r\\
0,& \text{otherwise}.
\end{cases}	
\end{equation}

\begin{figure*}[!t]
\centering
\includegraphics[width=17.4cm]{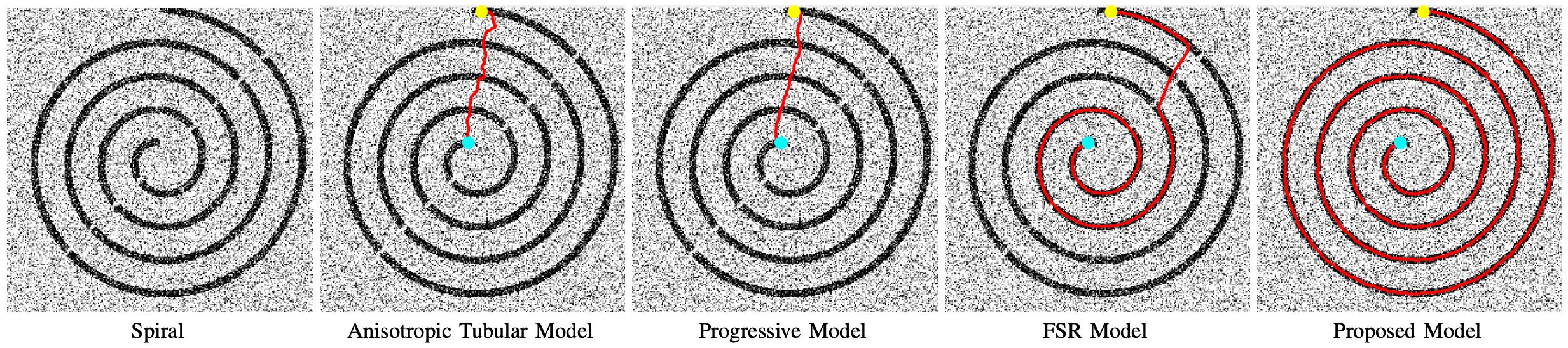}
\caption{Extraction of paths on a synthetic spiral image  interrupted by noise. \textbf{Column 1}: The spiral image with noise. \textbf{Columns 2-4}: The paths extracted from the anisotropic tubular minimal path model~\cite{benmansour2011tubular}, the progressive model~\cite{liao2018progressive}, the FSR minimal path model~\cite{duits2018optimal} and the proposed model, respectively}
\label{fig:Spiral}
\end{figure*}

The eigenvalues of $\Psi(\cdot)$, denoted by $\lambda_1(\cdot)$ and $\lambda_2(\cdot)$ with assumption $\lambda_1(\cdot)\leq\lambda_2(\cdot)$, can be used to characterize the appearance of tubular structure centerlines.  As in~\cite{law2008three}, one can compute a scalar-valued function $\zeta:\Omega\to\bR^+_0$, referred to as vessel score map,  such that $\zeta(x)$ is derived from $\lambda_1(x,\cdot)$ at the optimal scale
\begin{equation}
\label{eq:Vesselness}
\zeta(x)=\max\left\{\max_{r\in[R_{\rm min},R_{\rm max}]}\left\{-\frac{1}{r}\lambda_1(x,r)\right\},0\right\}.	
\end{equation}
By the eigenvalues $\lambda_1(\cdot)$, one can also derive an optimal scale map $\rho:\Omega\to[R_{\rm min},R_{\rm max}]$ as follows:
\begin{equation}
\rho(x)=\underset{r\in[R_{\rm min},R_{\rm max}]}{\arg\max}\left\{-\frac{1}{r}\lambda_1(x,r)\right\}.
\end{equation}
The score $\zeta(x)$ indicates the likelihood of a point $x$ belonging to a tubular structure. As in~\cite{law2008three,benmansour2011tubular}, the direction of a tubular structure at a point $x$  can be characterized by $\fv_2(x)$ which is the eigenvector of $\Psi(x,\rho(x))$ corresponding to the eigenvalue $\lambda_2(x,\rho(x))$. The tool of the orientation scores extends the confidence map $\Psi$ to a multi-orientation space. Denoting by $\psi_{\rm os}:\Omega\times\bS^1\to\bR^+$ the orientation scores associated to an image, we have
\begin{equation}
\label{eq:OS}
\psi_{\rm os}(x,\theta)=\max\{\langle\fn^\perp_{\theta},\,\Psi(x,r^*)\fn^\perp_{\theta}\rangle,0\},
\end{equation}
where $\fn_{\theta}^\perp=(-\sin\theta,\cos\theta)$ is a unit vector that is perpendicular to $\fn_\theta$.

Then the image data-driven function $\kC$ used in Eq.~\eqref{eq:CurvaMetrics} can be computed as 
\begin{equation}
\label{eq:OSCost}
\kC(\tx)=\exp\big(-\alpha\psi_{\rm os}(\tx)\big),
\end{equation}
where $\alpha\in\bR^+$ is a weighting parameter on the image data.

\subsection{Motivation}
In many applications, tubular structures involved in the images may consist of multiple tree structures. In image analysis applications, a fundamental task is to trace a path to delineate a curvilinear structure between two given points from the entire tree structures or from a complicated background. The classical tubular minimal path models~\cite{benmansour2011tubular,li2007vessels,cohen1997global} likely tend to travel along the regions with strong appearance features or directly pass through the background regions, yielding the shortcuts or short branches combination problems. 
Alternatively, exploiting minimal paths with curvature regularization for tracing tubular structures may reduce the risk of the above  problems in some extent. However, modeling  a tubular centerline as a globally optimal curvature-penalized geodesic path is not always suitable in practical applications. Moreover, the computation complexity for the curvature-penalized minimal path models are too high for real-time applications, since we expect the tracking time to be comparable to the user interaction time.  Examples for illustration of the shortcuts and short branches combination problems can be seen in~Figs.~\ref{fig:examples} and~\ref{fig:Spiral}. In Fig.~\ref{fig:examples}, the goal is to extract an artery vessel between two points from a retinal image patch. The results show that both the anisotropic tubular model and the FSR model suffer from the shortcuts and short branches combination problems, as depicted in Figs.~\ref{fig:examples}b and \ref{fig:examples}c.

Moreover, we make a test in Fig.~\ref{fig:Spiral} with a goal to  delineate a spiral from a synthetic image interrupted by noise. In this experiment, we evaluate the anisotropic tubular model~\cite{benmansour2011tubular}, the progressive model~\cite{liao2018progressive}, the FSR model and the proposed one. The results from the existing models suffer from the shortcuts problem due to the effects from the noise. 
In order to overcome these problems mentioned above,  we propose a new minimal path model for minimally interactive tubular structure tracing in conjunction with curvature-penalized minimal paths and trajectories grouping. The resulting path from the proposed model can avoid these problems as much as possible, as depicted in Fig.~\ref{fig:examples}d and in column $5$ of Fig.~\ref{fig:Spiral}.

\begin{figure*}[!t]
\centering
\includegraphics[width=17.5cm]{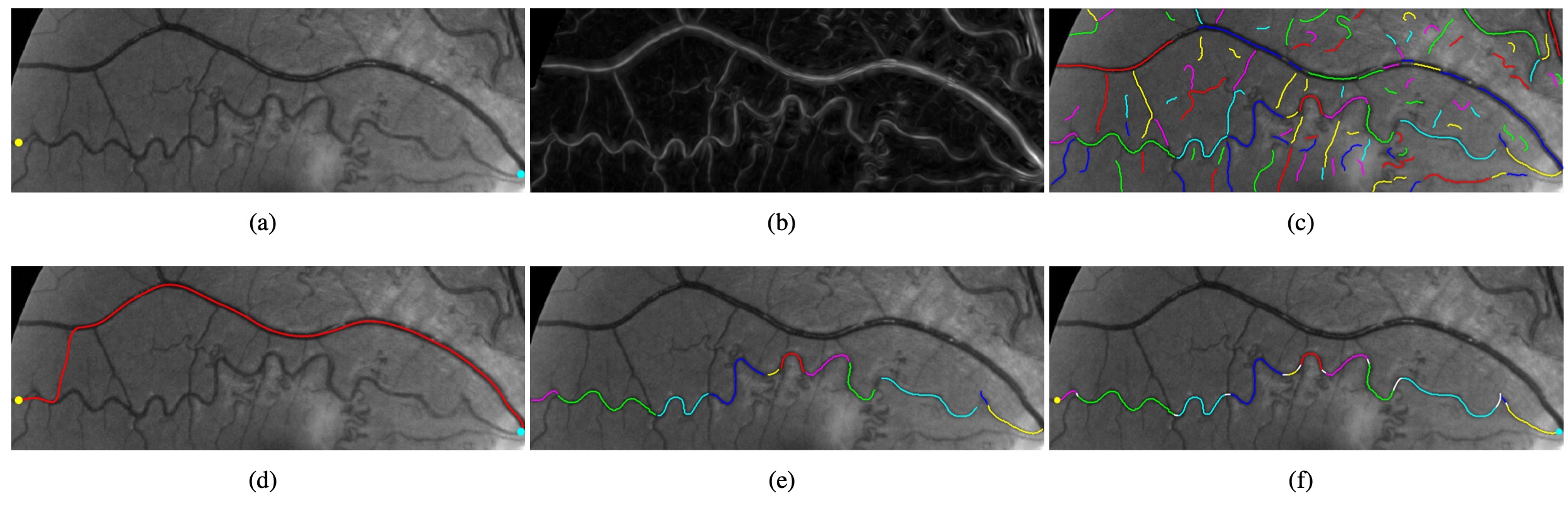}
\caption{The procedure of the proposed tubular structure tracing model. \textbf{a} A retinal image patch with vessels of high tortuosity. The dots represent the user-provided points.  \textbf{b}  Visualization for the vessel score map $\zeta$. \textbf{c} A set of trajectories superimposed on the image. \textbf{d} The geodesic path (indicated by red line) derived from the FSR minimal path model. \textbf{e} Grouped vessel trajectories. \textbf{f} The final path obtained by connecting the gaps between adjacent trajectories using bridging paths (white lines)}
\label{fig:procedure}	
\end{figure*}

\begin{figure*}[t]
\setlength{\fboxsep}{0pt}%
\centering
\subfigure[]{\fbox{\includegraphics[width=5.7cm]{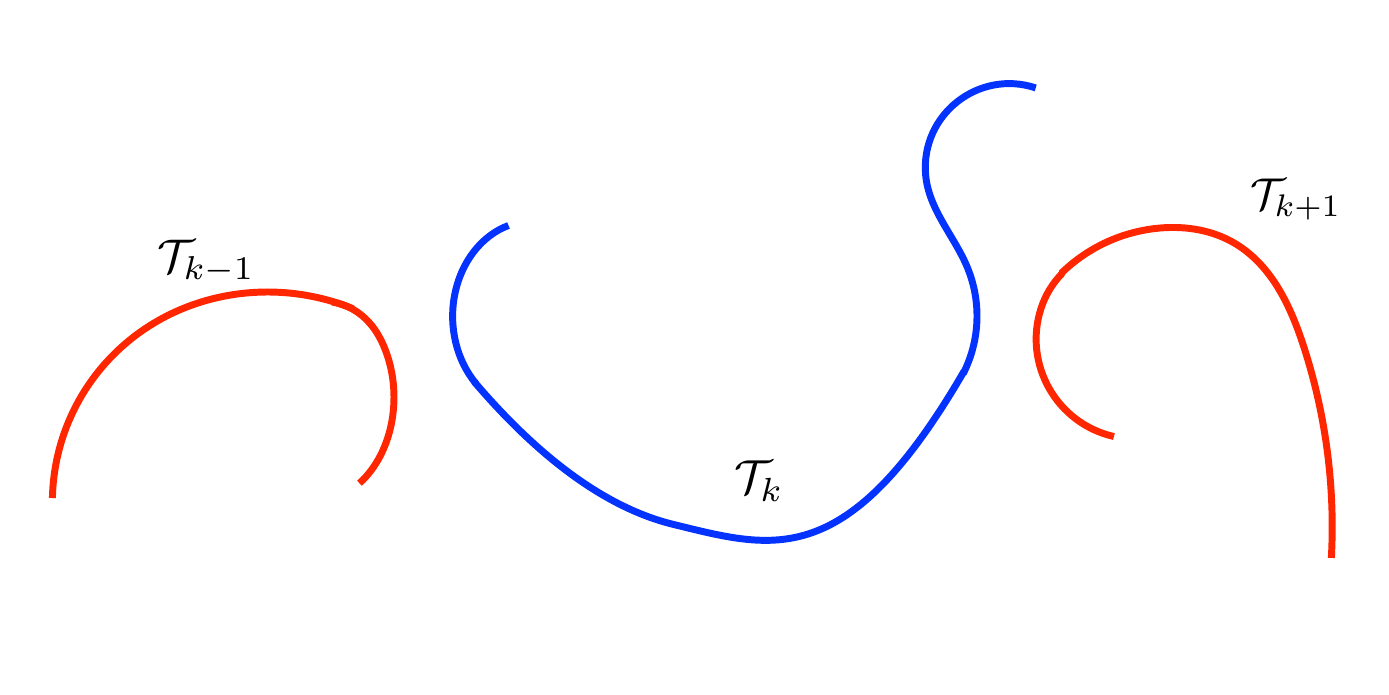}}}\,\subfigure[]{\fbox{\includegraphics[width=5.7cm]{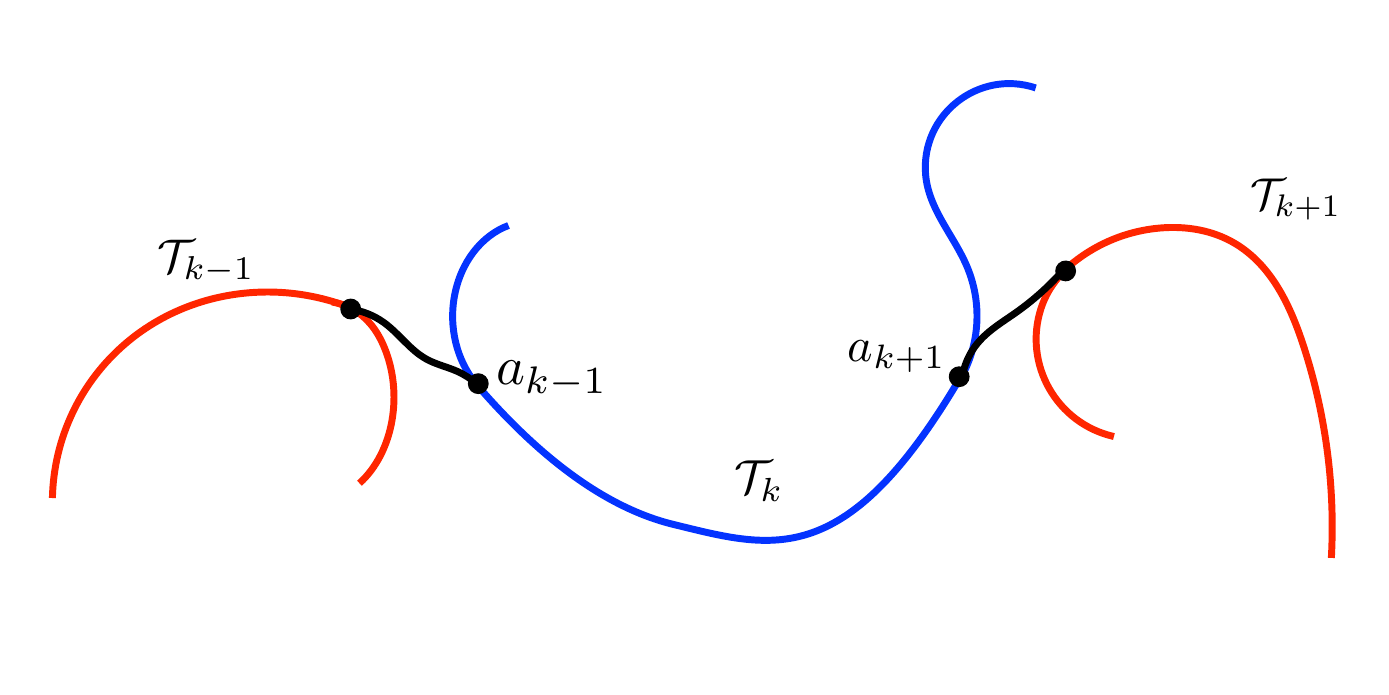}}}\,\subfigure[]{\fbox{\includegraphics[width=5.7cm]{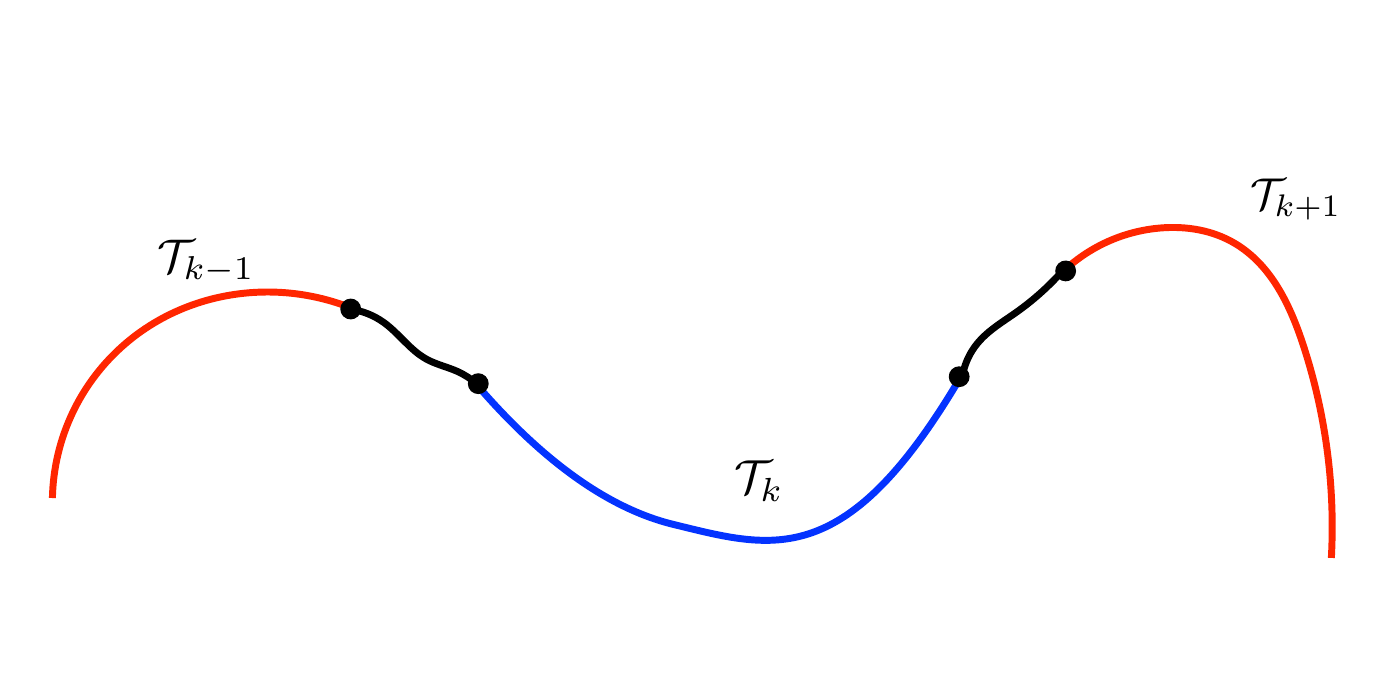}}}
\caption{Bridge the gaps between adjacent trajectories involved in an ordered sequence. \textbf{a} A trajectory $\cT_k$ (indicated by the blue line) adjacent to its neighbours $\cT_{k-1}$ and $\cT_{k+1}$ indicated by red lines. \textbf{b} Bridging paths indicated by green lines. The black dots lying at the trajectory $\cT_k$ are the points $a_{k-1}$ and $a_{k+1}$, see text. \textbf{c} The truncated  trajectories (red lines) and curvature-penalized bridge paths (black lines).}
\label{fig:Concatenation}
\end{figure*}

\section{Trajectory Grouping for Tracing Tubular Structures}
\label{sec:PG}
\subsection{Disjoint Trajectories as Rough Tubularity Descriptor}
\label{subsec_Trajectory}
Basically, tubularity segmentation is regarded as a way of classifying all points into either tubular structures or background. The segmentation is usually represented  by a binary mapping, which can be generated by many tubular structure segmentation approaches as reviewed in literature~\cite{moccia2018blood,lesage2009review}. In this paper, we implement a simple method to segment the tubular structures  by directly thresholding a vessel score map $\zeta$, see Eq.~\eqref{eq:Vesselness}. An example for the visualization of the score map $\zeta$ that is derived from a retinal image patch can be seen in Fig.~\ref{fig:procedure}b.

 Following that we apply the morphological filters on tubularity segmentation to get the skeleton structure of entire tubular structure network.  In this paper, we suppose that each trajectory is a connected set involving skeleton points. In order to obtain disjoint trajectories, we remove all the branch points from the computed tubular skeleton structures, where a branch point is a skeleton point connecting at least three trajectories.  We illustrate an example for the disjoint trajectories in Fig.~\ref{fig:procedure}c, where each trajectory is tagged by different colors. Moreover, in order to reduce the computation complexity, we remove the trajectories for which the length (in grid points) is lower than a given thresholding value.

\subsection{Graph-based Shortest Path}
\label{subsec_GraphSP}
The proposed model for tracing tubular structure centerlines  is established on a graph $\cG=(V,E)$, where $V$ represents the  set of nodes and $E$ stands for the set of edges between connected nodes. We denote by $e_{ij}\in E$ an edge joining two adjacent nodes $\vartheta_i$, $\vartheta_j\in V$. Each edge $e_{ij}$ is assigned a weight $\omega_{ij}\in\bR^+_0$. For the sake of simplicity, we leverage  the convention that the weight $\omega_{ij}=+\infty$ implies the node $\vartheta_i$ is disconnected to $\vartheta_j$. 
In tubular structure tracking,  one may expect to get the same tracking result by choosing either of the two prescribed points as the source point and taking the remaining one as end point. For this purpose, we focus on the \emph{indirect} graph, which means that for each pair of edges $e_{ij}$ and  $e_{ji}$,  one has $\omega_{ij}=\omega_{ji}$. 

As in~\cite{wang2013interactive}, the graph $\cG$ is constructed  by associating a node $\vartheta_i\in V$ to a trajectory $\cT_i$. The edge set $E$ characterizes the connection between two trajectories $\cT_i$ and $\cT_j$ for any $i\neq j$. For a fixed trajectory $\cT_i$, the joint trajectories $\cT_j$ can be detected by a tubular neighbourhood $M_i\subset\Omega$ surrounding $\cT_i$. A possible way is to perform front propagation associated to a suitable metric emanating from  $\cT_i$ such that $M_i$ can be identified by thresholding the propagated geodesic distance. However, this method may increase the computation complexity of the entire algorithm. Alternatively, we make use of the method proposed in~\cite{wang2013interactive} which uses Euclidean distance. Specifically, the trajectory $\cT_i$ is first prolonged from its two endpoints along the respective tangents, in order to generate an prolonged trajectory $T_i$. Then we build a regular tubular neighbourhood $M_i$ for the extended trajectory by
\begin{equation}
M_i=\left\{x\in\Omega\, |\, \min_{y\in T_i}\|x-y\|<\tau\right\},
\end{equation}
where $\tau$ is a given thresholding value. A trajectory $\cT_j$ is said to be adjacent to the fixed $\cT_i$ for any  $j\neq i$, if  the prolongated trajectories $T_j$ satisfies 
\begin{equation}
M_i\cap T_j\neq\emptyset.	
\end{equation}

Given two vertices, optimizing  the  graph $\cG$ by Dijkstra's algorithm~\cite{dijkstra1959note} yields a shortest path that consists of a series of ordered trajectories. 
Once the sets $V$ and $E$  for the graph $\cG$ are built, one can point out that the weight $\omega_{ij}$ dominates the tracing results. The weight $\omega_{ij}$ measures the cost of bridging the gap between a pair of adjacent trajectories $\cT_i$ and $\cT_j$. In contrast to~\cite{wang2013interactive} which exploits straight segments, we complete the gap between a pair of adjacent trajectories by curvature-penalized geodesic paths.

\subsection{Computing the Weights $\omega_{ij}$}
The generation of a set of orientation-lifted trajectories constitutes the first step in the proposed model. 
Basically, a trajectory $\cT_i\subset\Omega$ likely passes through the centerline of a tubular structure. Thus, a point $x\in \cT_i$ can be assigned to a pair of  orientations $\theta_x\in[0,\pi]$ and $\theta_x+\pi$ which characterize the directions of the centerline at $x$. Such an orientation $\theta_x$ can be computed via the orientation scores $\psi_{\rm os}$ as follows
\begin{equation}
\label{eq:optimalOrien}
\theta^*_x=\underset{\theta\in[0,\pi]}{\arg\max}~\psi_{\rm os}(x,\theta).
\end{equation}
Accordingly, a trajectory $\cT_i$ can be lifted to the space $\tilde\Omega$, leading to a new subset $\widetilde\cT_i\subset\tilde\Omega$ as a union of two sets
\begin{equation}
\widetilde\cT_i=\{(x,\theta^*_x)\mid\forall x\in\cT_i\}\cup\{(x,\theta^*_x+\pi)\mid\forall x\in\cT_i\}.	
\end{equation}
We say that two orientation-lifted trajectories $\widetilde\cT_i$ and $\widetilde\cT_j$ for any $i\neq j$ are adjacent if their physical projections  $\cT_i$ and $\cT_j$ are adjacent.

In the proposed model, we expect to choose a set of trajectories $\cT_i$ to constitute  a shortest path which delineates the target tubular structure, providing that two vertices are given. In principle, the weights $\omega_{ij}$ should be as low as possible if two adjacent trajectories $\cT_i$ and $\cT_j$ lie at the same tubular structure. For a fixed trajectory $\cT_i$ lying in the target structure, the construction for the weights must allow to differentiate between a trajectory that belongs to  the same structure with $\cT_i$ and one belongs to another structure.  
In many scenarios, tubular structures appear to be locally smooth. Thus it is reasonable to estimate the edge weights $\omega_{ij}$ using curvature-penalized geodesic distance. The smoothness property of the curvature-penalized minimal paths agrees with the requirement for connecting the gaps between two adjacent vessel segments.

For a fixed trajectory $\widetilde\cT_i$, we consider a geodesic distance map $\cU_i$ with respect to $\widetilde\cT_i$ such that
\begin{align}
\label{eq_DistToSet}
\cU_i(\tx)=\inf\Big\{\cL_\epsilon(\tilde\gamma)| &\tilde\gamma\in\Lip([0,1],\tilde\Omega),\nonumber\\
&\tilde\gamma(0)\in\widetilde\cT_i,\tilde\gamma(1)=\tx\Big\},
\end{align}
where $\cL_\epsilon(\tilde\gamma)$ is the weighted curve length of $\tilde\gamma$ associated to the metric $\cF_{\epsilon}$, see Eq.~\eqref{eq_CurveLength}.
As discussed in Section~\ref{sec:MPs}, the geodesic distance map $\cU_i$ admits the eikonal PDE~\eqref{eq:EikonalPDE} with a boundary condition $\cU_i(\tx)=0$ for any point $\tx\in\widetilde\cT_i$.

By the definition~\eqref{eq_DistToSet}, we can estimate a distance value $\cD_{i,j}$ that is  the minimal weighted curve length between the fixed trajectory $\widetilde\cT_i$ and a neighbouring one $\widetilde\cT_j$ 
\begin{equation}
\cD_{i,j}=\min_{\tx\in\widetilde\cT_j}\cU_i(\tx).
\end{equation}

Once the distance map $\cU_i$ is computed,  a geodesic path $\cC_{i,j}\in\Lip([0,1],\tilde\Omega)$ such that $\cC_{i,j}(0)\in\widetilde\cT_i$  and $\cC_{i,j}(1)\in\widetilde\cT_j$ can be tracked by performing the gradient descent ODE on the distance map $\cU_i$, see Eq.~\eqref{eq:ODE}.

The minimal length $\cD_{i,j}$ associated to the metric $\cF_\epsilon$ encodes the image data-driven function $\kC$ defined in Eq.~\eqref{eq:CurvaMetrics}, due to the use of the metric $\cF_{\epsilon}$. Accordingly, the length $\cD_{i,j}$ encodes the tubular appearance features. This may introduce bias to the cost of connecting the trajectory $\cT_i$ to any neighbour $\cT_j$, especially when the target structures  weaker than its neighbouring ones. In order to remove the effect from the image data, we take into account a new weighted length $\kD_{i,j}$ measured along the geodesic path $\cC_{i,j}$ derived from the distance map $\cU_i$. Let us denote $\cC_{i,j}:=(\rC_{i,j},\tau_{i,j})$ where $\rC_{i,j}(u)$ indicates the spatial positions of the orientation-lifted geodesic path $\cC_{i,j}$, and the parametric function $\tau_{i,j}(u)$ determines the tangent $\rC^\prime_{i,j}$, see Eq.~\eqref{eq_TurningAnkle}. We expect that the quantity $\kD_{i,j}$ is explicitly dependent to the curvature of $\rC_{i,j}$, but independent to the image data $\kC$. Thus the quantity $\kD_{i,j}$ can be estimated by
\begin{align}
\kD_{i,j}&=\int_0^1 \sqrt{\|\rC^\prime_{i,j}(u)\|^2+\beta^2\tau^\prime_{i,j}(u)^2}du\nonumber\\
\label{eq:data_ind}	
&=\int_0^1\sqrt{1+\beta^2\kappa_{i,j}(u)^2}\,\|\rC_{i,j}^\prime(u)\|du,
\end{align}
where $\kappa_{i,j}$ denotes the curvature of the physical projection $\rC_{i,j}$. The parameter $\beta\in\bR^+$ is a constant controlling the importance of the curvature.
Note that the estimation of $\kD_{i,j}$ by Eq.~\eqref{eq:data_ind} allows to give more importance to the curvature penalization by setting a larger value to parameter $\beta$ than the one used for defining $\cD_{i,j}$. Such a quantity $\kD_{i,j}$ can be computed directly through the path $\cC_{i,j}$. In this paper, we also consider an alternative way for estimating the quantity $\kD_{i,j}$ relies on a new map $\cE_i:\tilde\Omega\to\bR^+_0$. For each point $\tx\in\tilde\Omega$, we can obtain a geodesic path $\cC=(\rC,\tau)$ such that $\cC(0)\in\widetilde\cT_i$ and $\cC(1)=\tx$. Similar to  Eq.~\eqref{eq:data_ind}, we set
\begin{equation}
\cE_i(\tx)=	\int_0^1\sqrt{1+\beta^2\kappa(u)^2}\,\|\rC^\prime(u)\|du,
\end{equation}
where $\kappa$ represents the curvature of the physical projection $\rC$. We denote by $\tx_i^*$ an optimal point such that $\cD_{i,j}=\cU_i(\tx_i^*)$, yielding that $\kD_{i,j}=\cE_i(\tx_i^*)$. Note that both of the maps $\cU_i$ and $\cE_i$ can be computed simultaneously,  without tracking the geodesic paths $\cC$, as discussed in~\cite{deschamps2001fast,chen2021elastica}.
In the following experiments, we apply  the map $\cE_i$ for  the estimation of the value $\kD_{i,j}$.
 
Similar to the computation of the geodesic path $\cC_{i,j}$, we can generate a different geodesic path $\cC_{j,i}=(\rC_{j,i},\tau_{j,i})$ traveling from the trajectory $\cT_j$ to $\cT_i$ subject to $\cC_{j,i}(0)\in\widetilde\cT_j$ and $\cC_{j,i}(1)\in\widetilde\cT_i$. By Eq.~\eqref{eq:data_ind}, we can obtain a weighted curve length $\kD_{j,i}$ associated to $\cC_{j,i}$. 

As a consequence, we define the weight $\omega_{ij}$ for the edge $e_{ij}$ as follows
\begin{equation}
\label{eq:edgeWeight}
\omega_{ij}=\min\{\kD_{ij},\,\kD_{ji}\}.
\end{equation}
The definition in Eq.~\eqref{eq:edgeWeight} imposes $\omega_{ij}=\omega_{ji}$, where $\omega_{ij}$, $\omega_{ji}$ are the respective weights for the edges of $e_{ij}$ and $e_{ji}$. Following that, the path completing the gap between $\cT_i$ and $\cT_j$ can be chosen as the minimal one between $\cC_{i,j}$ and $\cC_{j,i}$ in the sense of the weighted curve length formulated in Eq.~\eqref{eq:data_ind}. In the following, the geodesic paths for connecting the gaps are referred to as \emph{bridge paths}. 

\begin{algorithm}
\caption{Numerical Implementation} 
\label{alg_FM} 
\renewcommand{\algorithmicrequire}{\textbf{Input:}}
\renewcommand{\algorithmicensure}{\textbf{Output:}}
\renewcommand{\algorithmiclastcon}{\textbf{Initialization:}}
\begin{algorithmic}[1]
\Require A pair of orientation-lifted trajectories $\widetilde\cT_i,\,\widetilde\cT_j$, and the neighbourhood system $\mathcal{N}$.
 \lastcon  
\dotdef $\forall \tx\in\widetilde\cT_i$, set $\cU_i(\tx)\leftarrow 0$, $\cE_i(\tx)\leftarrow 0$, $\mathfrak{L}(\tx)\leftarrow \mathrm{Trial}$. 
\dotdef $\forall \tx\in\tilde\Omega\backslash\widetilde\cT_i$, set $\cU_{i}(\tx)\leftarrow \infty$, $\cE_i(\tx)\leftarrow \infty$, $\mathfrak{L}(\tx)\leftarrow \mathrm{Far}$.
\While{$\mathfrak{L}(\tx)\neq\mathrm{Accepted}$ for any point $\tx\in\widetilde\cT_j$}
\State Find $\tx_{\mathrm{min}}$ that is the $\mathrm{Trial}$ point  minimizing $\cU_i$.
\State Set $\tx^*\gets\tx_{\mathrm{min}}$
\State Set $\mathfrak{L}(\tx_{\mathrm{min}})\leftarrow \mathrm{Accepted}$.
\State Update the value $\cE_i(\tx_{\mathrm{min}})$.
\label{algLine_E}
\ForAll {$\tp$ such that $\tx_{\mathrm{min}}\in\mathcal{N}(\tp)$}
\If{$\mathfrak{L}(\tp)\neq \mathrm{Accepted}$}
\State Set $\mathfrak{L}(\tp)\leftarrow \mathrm{Trial}$.
\State Update the distance $\cU_{i}(\tp)$. 
\label{algline_Distance}
\EndIf  
\EndFor 
\EndWhile 
\State Track the geodesic path $\cC_{i,j}$ via Eq.~\eqref{eq:ODE}.
\State Set $\kD_{i,j}=\cE_i(\tx^*)$.
\end{algorithmic}	
\end{algorithm}

\subsection{Implementation Consideration} 
\subsubsection{Fast marching Implementation}
The numerical computation for the maps $\cU_{i}$ and $\cE_i$, indexed by $i$, can be efficiently implemented using state-of-the-art Hamiltonian fast marching algorithm (HFM)~\cite{mirebeau2018fast}, as presented in Algorithm~\ref{alg_FM}. The HFM is performed on a regular grid $\bM=(\Omega\cap h_1\bZ^2)\times(\bS^1\cap h_2\bZ)$ where $h_1$ and $h_2$ represent the discretization scales. In the experiments, we set $h_1=1$ and $h_2=2\pi/N_\theta$, where $N_\theta$ is the number of discretized orientations that is fixed as $N_\theta=60$.

For each trajectory $\widetilde\cT_i$ and a set of neighbouring trajectories $\{\widetilde\cT_j\}_j$, we perform the HFM by taking all the grid points involved in the set $\widetilde\cT_i\cap\bM$ as the source points for initializing the maps $\cU_{i}$ and $\cE_i$, and terminate  the HFM whenever a grid point in the set $\widetilde\cT_j\cap\bM$ is reached. The estimation of the map $\cE_i$ is carried out in an accumulation manner~\cite{deschamps2001fast,chen2021elastica}. During the distance estimation scheme, the geodesic distance map $\cD_{i,j}$ is updated by solving the discretization of the eikonal PDE with a Hamiltonian form~\cite{mirebeau2018fast}, as in Line~\ref{algline_Distance} of Algorithm~\ref{alg_FM}. Such a scheme can also update the geodesic flows for estimating the map $\cE_i$, see Line~\ref{algLine_E} of Algorithm~\ref{alg_FM}. 

\begin{figure*}[t]
\centering
\includegraphics[width=17.5cm]{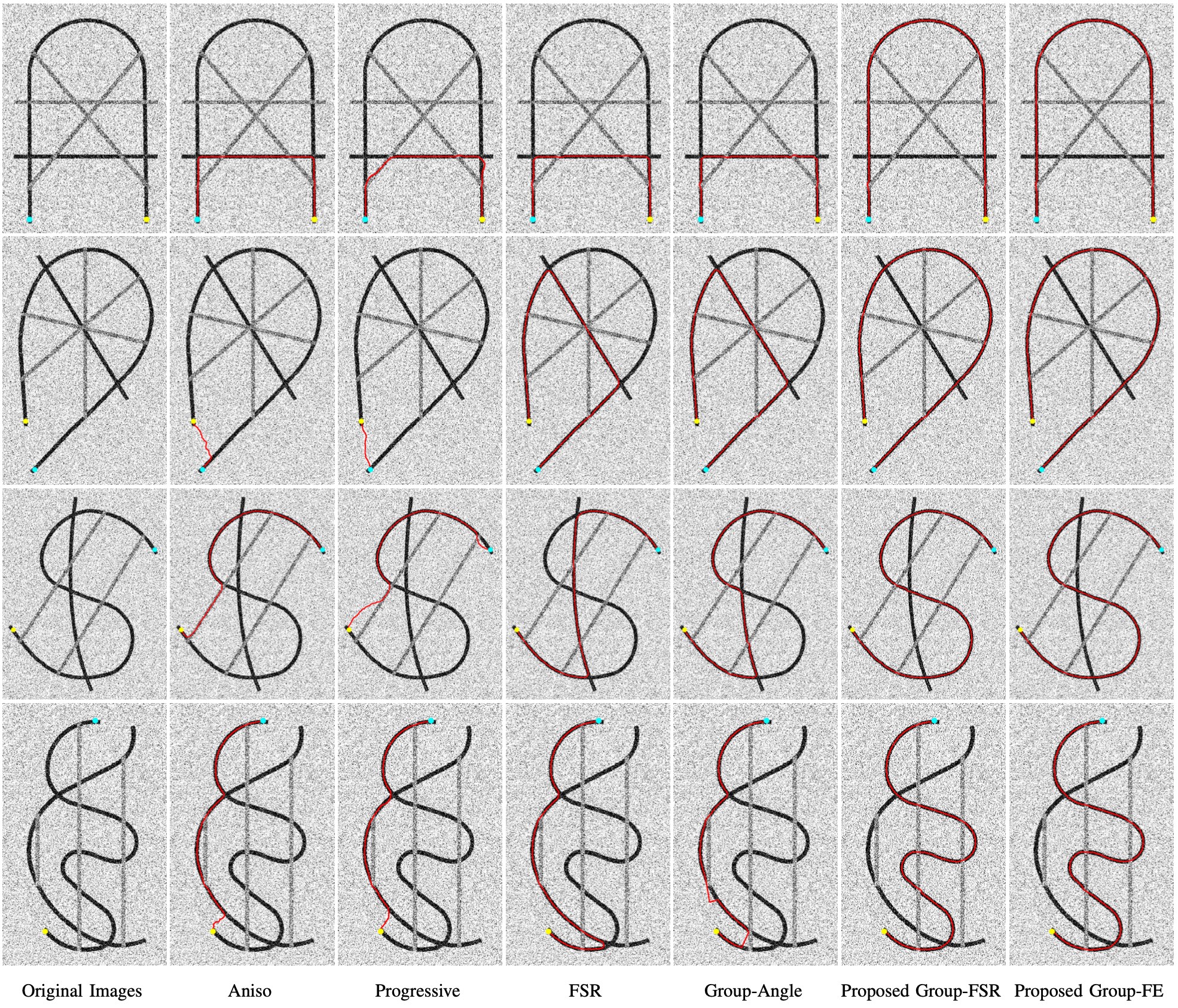}
\caption{Qualitative comparison between different models on synthetic images. The cyan and yellow dots indicate the source and end points, respectively.  \textbf{Column} 1: The original images. \textbf{Columns} 2-7: The red lines represent the obtained paths from each model. }
\label{fig:SyntheticImages}
\end{figure*}

\subsubsection{Recovering optimal paths}
The shortest path $\rT$ derived from the Dijkstra's algorithm~\cite{dijkstra1959note} consists of a set of $K$ ordered nodes $\cT_k$ with $1\leq k \leq K$ and $K\geq3$, where each node is a trajectory. In Fig.~\ref{fig:procedure}e, we show an example of such a series of ordered trajectories $\cT_k$.  In order to build the final connected path $\rT^*$, one has to complete the gaps between each pair of adjacent trajectories involved in the shortest path $\rT$~\cite{wang2013interactive}. In other words, the target path $\rT^*$ can be regarded as a sequence of trajectories and bridge paths. Note that the first and last nodes in $\rT$, i.e. $\cT_1$ and $\cT_K$, are the source and end vertices provided by the user. Among the remaining trajectories of $\rT$ , we take a trajectory $\cT_k$ ($k>1$) as example  to show how to bridge the corresponding gaps with its neighbours  $\cT_{k+1}$ and $\cT_{k-1}$.  The bridge paths $\cC_{k,k+1}$ and $\cC_{k,k-1}$ between each neighbouring trajectory and $\cT_k$ will intersect $\cT_k$ at two points $a_{k-1}$ and $a_{k+1}$, as depicted in Fig.~\ref{fig:Concatenation}. Only the portion of $\cT_k$ between points $a_{k-1}$ and $a_{k+1}$ is involved in $\rT^*$, which is referred to as $\cT^*_k$. The final path is built as the union of all the truncated trajectories $\cT^*_k$ and the bridge paths, as depicted in Fig.~\ref{fig:procedure}f and Fig.~\ref{fig:Concatenation}c. 

\subsubsection{Computation Complexity}
The computation cost for the proposed model can be divided into two parts. The first one lies at the construction of the graph, which mainly consists of  the computation for the edge weights. This step indeed cost very long execution time. Fortunately, this process can be done offline such that the user do not have to wait online in this stage~\cite{wang2013interactive}. Also,  the parallel computing scheme can greatly speed up the computation. 

For the Dijkstra's shortest path algorithm, the computation complexity is $\mathcal{O}(N\log N)$ where $N$ is the total number of trajectories. Since $N$ is much less than the number of grid points of an image, the proposed tubular structure tracking method is able to yield shortest paths in a real-time manner,  i.e. comparable to the user interaction.

\section{Experimental Results} 
\label{sec:Exp}
\subsection{Configuration}
We conduct the numerical experiments with both qualitative and quantitative comparison to  the progressive minimal path model with bending constraint  (Progressive)~\cite{liao2018progressive}, the anisotropic tubular minimal path model (Aniso)~\cite{benmansour2011tubular}, the curvature-penalized minimal path model with a FSR metric~\cite{duits2018optimal}, the grouping method with Euclidean distance and angles (Group-Angle)~\cite{wang2013interactive} on both synthetic and real images. We refer to the proposed model as Group-FE (resp. Group-FSR) if the edge weights $\omega_{ij}$ of the graph $\cG$ is estimated by the FE metric (resp. FSR metric).

\begin{figure*}[t]
\centering
\includegraphics[width=17.5cm]{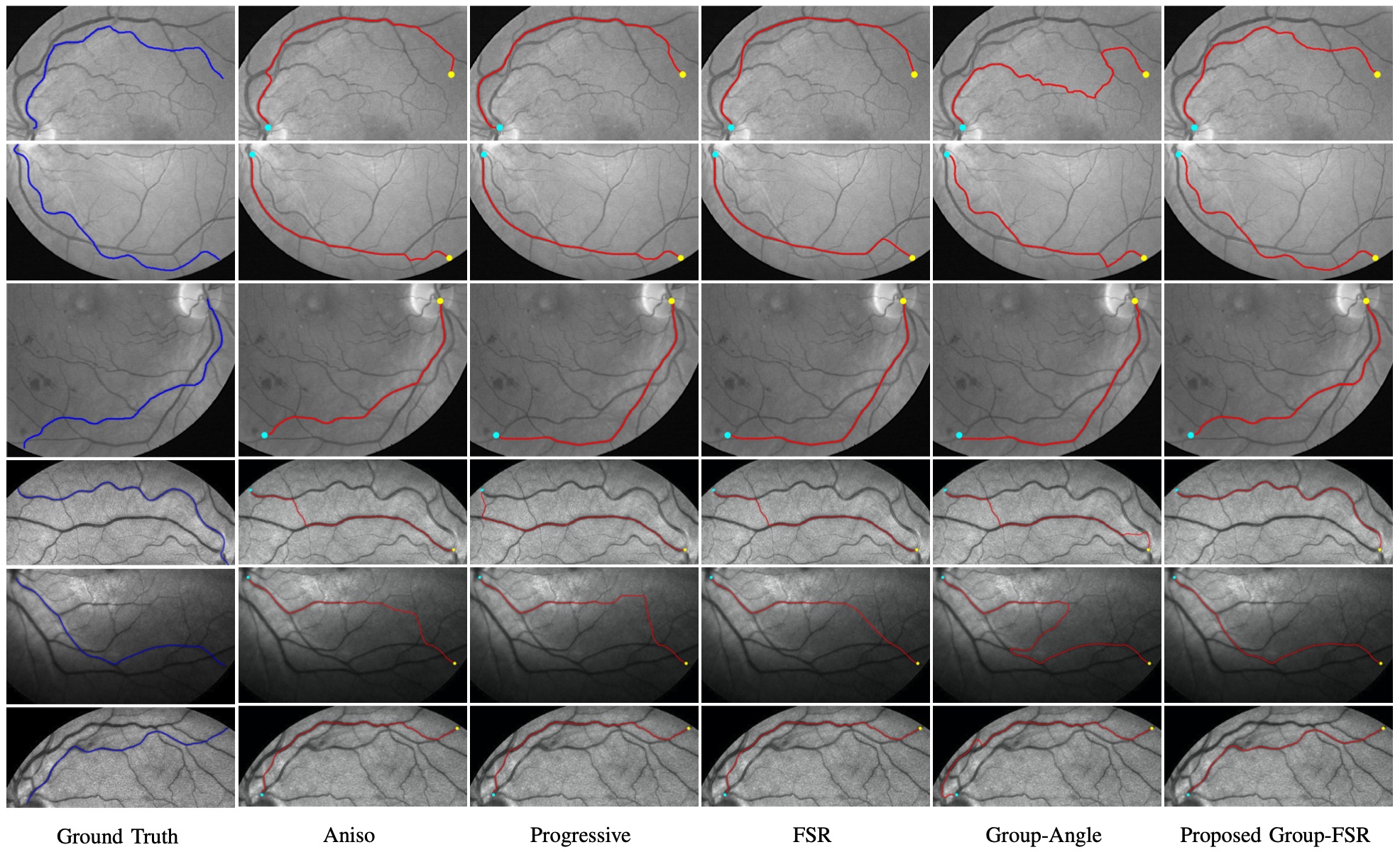}
\caption{Qualitative comparison between different models on retinal image patches. \textbf{Column} 1: The blue lines indicate the target tubular structure. \textbf{Columns} 2-6: The cyan and yellow dots indicate the source and end points, respectively. The resulting paths (red lines) derived from the Aniso model, the Progressive model, the FSR model, the Group-Angle model and the proposed Group-FSR model, respectively.}
\label{fig:QuaRetina}
\end{figure*}

\begin{figure*}[t]
\centering
\includegraphics[width=17.5cm]{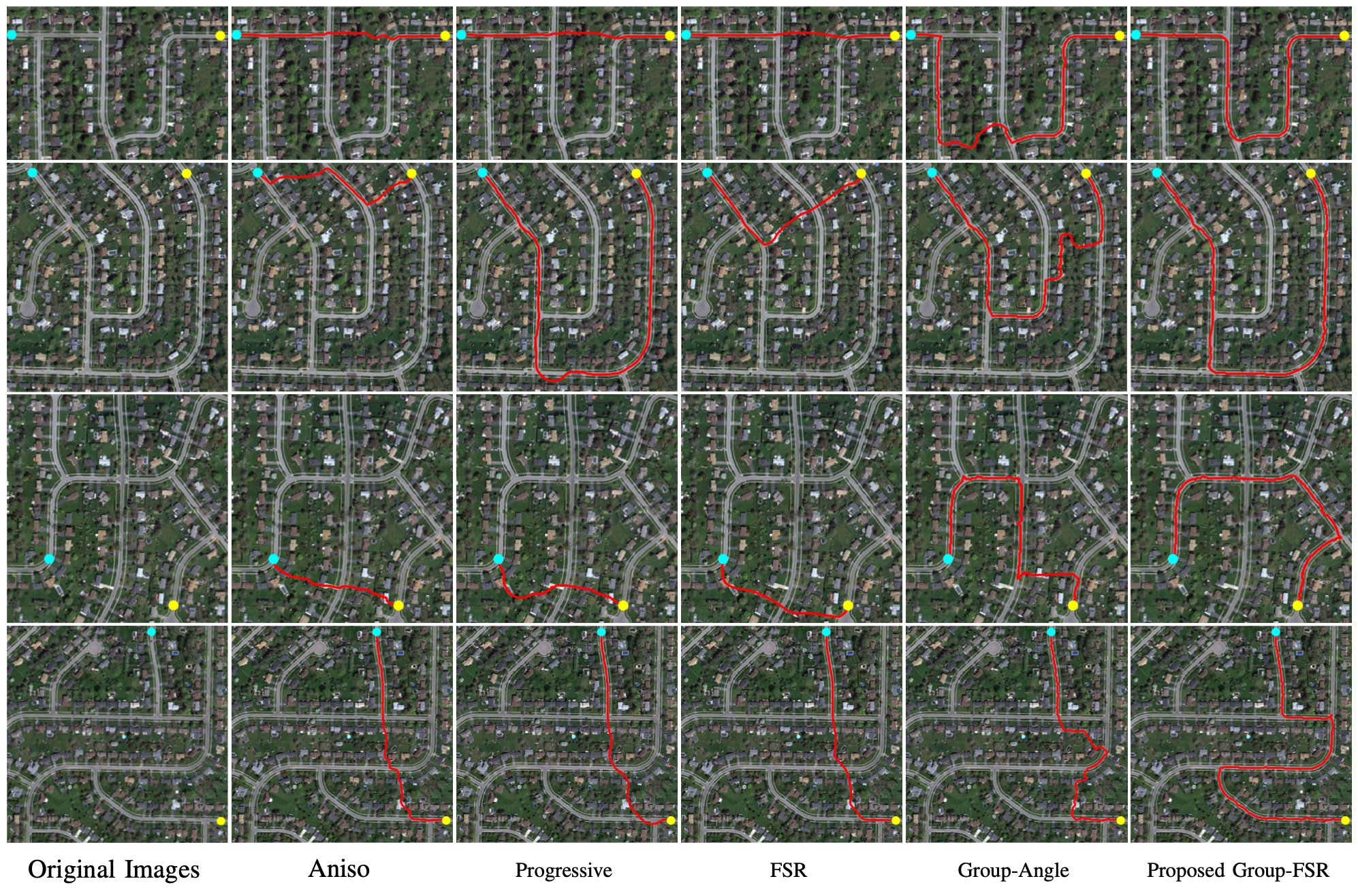}
\caption{Qualitative comparison between different models on road image patches. The dots indicate the user-provided points and the red lines represent the obtained paths. \textbf{Column} 1: The original images. \textbf{Columns} 2-6: Results from the Aniso, Progressive, FSR, Group-Angle and the proposed Group-FSR models, respectively.}
\label{fig:QuaRoads}
\end{figure*}

\begin{figure*}[t]
\centering
\includegraphics[width=17.5cm]{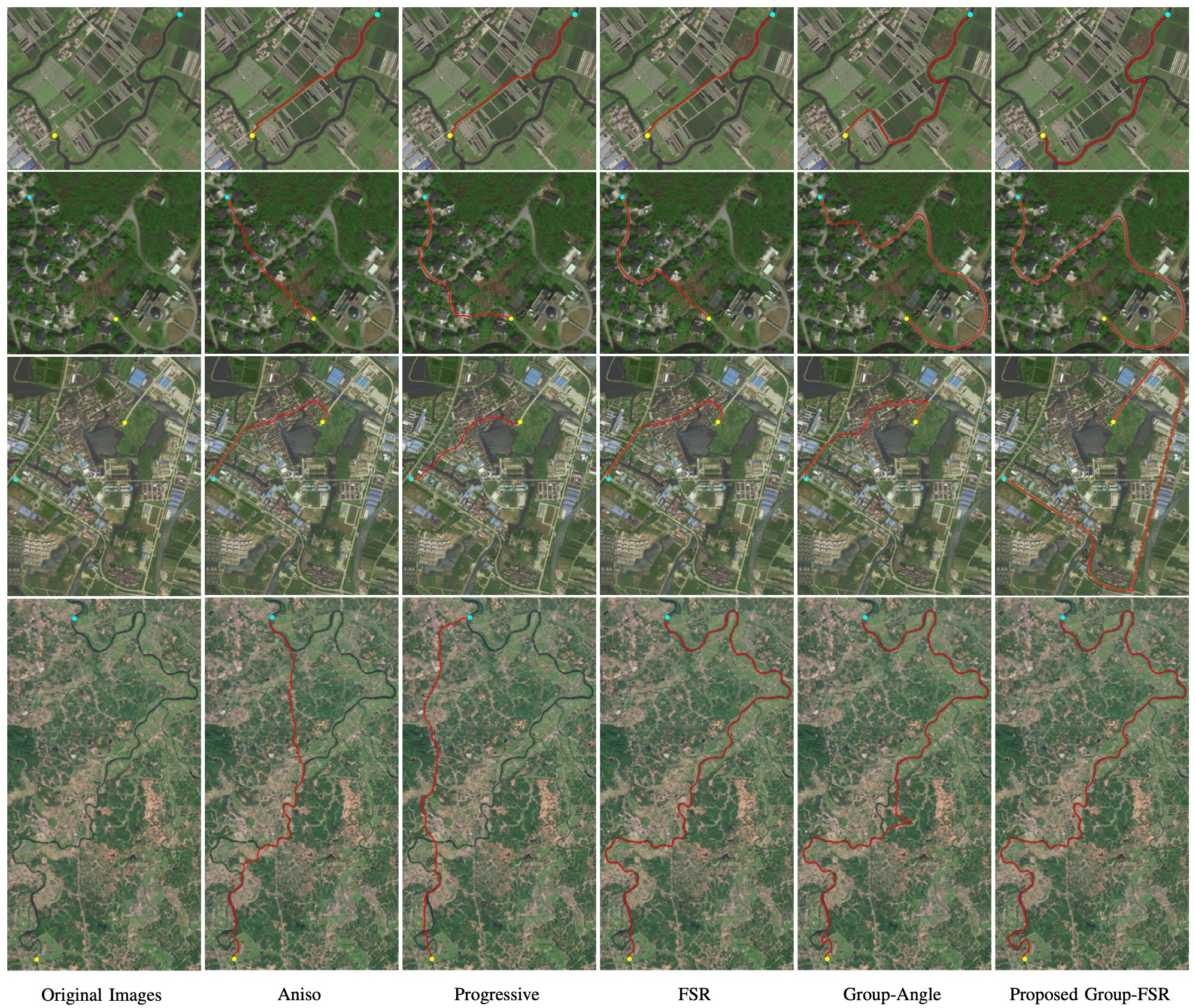}
\caption{Qualitative comparison between different models on road and river image patches. The dots indicate the user-provided points and the red lines represent the obtained paths. \textbf{Column} 1: The original images. \textbf{Columns} 2-6: Results from the Aniso, Progressive, FSR, Group-Angle and Group-FSR models, respectively.}
\label{fig:QuaRiverRoads}
\end{figure*}

The Progressive minimal path model invokes a dynamic isotropic metric by taking into account nonlocal path features, which introduces the bending constraint into the fast marching fronts propagation scheme, in order to avoid the shortcuts and short branches combination problems. The path feature is computed using a backtracked truncated geodesic path whose length is fixed as $10$ grid points. The thresholding value that defines the maximally admissible bending degree is set to $0.9$, see~\cite{liao2018progressive} for more details.  In the following experiments, we use the same model as~\cite{benmansour2011tubular} to construct the anisotropic Riemannian metric for the Aniso model. The anisotropic ratio for this metric is fixed to $10$ for all the tests.

The FSR and FE minimal path models  descried in Section.~\ref{sec:MPs} are also considered in the experiments. For the FSR and FE metrics, we fix the parameter $\epsilon=0.1$ to achieve good balance between the accuracy and the computation complexity. The parameters $\alpha$ and $\beta$ are weighted parameters which control the importance of the image data and the curvature penalization, respectively. We set $\alpha=5$ and $\beta=20$ for all the tests. In addition, both metrics are also used to estimate the edge weights $\omega_{ij}$ for the proposed Group-FE and Group-FSR models. 
For the Group-Angle model, the same edge set $V$ as the proposed  Group-FE model is adopted to construct the graph.
We exploit the identical method as in ~\cite{wang2013interactive} for the estimation of the edge weights $\omega_{ij}$.

In the OOF filter, the Gaussian kernel $G_\sigma$ with the standard deviation  $\sigma$ is used to slightly smooth of the input image. The interval $[R_{\rm min},R_{\rm max}]$ is the range of the possible radii. We set $\sigma=1.5$, $R_{\rm min}=1$, and $R_{\rm max}=8$. For numerical consideration, we normalize  $\psi_{\rm os}$ to the range $[0,1]$.

Finally, all the experiments are performed on a standard 6-core Intel Core i7 of 3.2GHz architecture with 64Gb RAM.

\subsection{Qualitative Comparison Results}
\label{sec:QualitativeExp}
In this work, we compare the proposed model with the state-of-the-art minimal path-based tracing algorithms.  

In Fig.~\ref{fig:SyntheticImages}, we show the tubular centerline tracing results on our synthetic images. The average size of the four images is $474\times321$ grid points. The gray levels of these synthetic images are normalized to the range $[0,1]$ interrupted by additive Gaussian noise of normalized standard derivation $\sigma_n=0.15$. In each image, there exist two tubular structures of low gray levels, along which the directions vary smoothly. Our goal is to extract the one between two given points. In rows $1$ and $2$, one can see that the source and end points are close to one another and the target structure has a long Euclidean length. In rows $3$ and $4$, the target structures have strong tortuosity appearance. In Fig.~\ref{fig:SyntheticImages}, columns $2$ to $5$ illustrate the results from the Aniso, Progressive, FSR and Group-Angle models, where we have observed the occurrence of the shortcuts and short branches combination problems. While the paths generated by the proposed Group-FSR and Group-FE models can correctly trace the objective vessels, as depicted in columns $6$ and $7$columns $5$ and $6$, due to the use of smooth geodesic paths for the connection of gaps between trajectories.

 In Fig.~\ref{fig:QuaRetina}, we qualitatively compare the proposed Group-FSR model\footnote{We observed that tracing results from the proposed Group-FE model are almost identical to those from Group-FSR model. For better visualization, we only show the results from the Group-FSR model.} to the Aniso, Progressive, FSR and Group-Angle models on from the DRIVE~\cite{staal2004ridge} and IOSTAR~\cite{zhang2016robust} datasets. The original retinal images are RGB color images, where we only use the green channel of each test image~\cite{zhang2016robust} for the computation of the cost function $\mathfrak{E}$, see Eq.~\eqref{eq:OSCost}. The goal is to trace an artery vessel between the given points indicated by dots. The corresponding ground truth is depicted in column $1$. We note that each model is performed in the entire retinal image and we only show the image patch in gray level containing the target vessel for well visualization.  
 In retinal images, the main challenge is that artery vessels appear weaker and are often close to or even cross over veins of strong visibility. In this experiment, some portions of the target vessels in general appear to be strongly tortuous. This may introduce difficulties to the Progressive and FSR models. Moreover, the strong neighbours of the target vessels will highly affect the results of the Aniso model. Finally, using straight segments to link adjacent trajectories for vessels of high length may accumulate the approximation errors. In columns $2$ to $5$ of Fig.~\ref{fig:QuaRetina}, we can see that the artery vessel tracing results from the existing models again suffer from the shortcuts and short branches combination problems.  In contrast, the proposed Group-FSR model passed by the correct way, see column $6$.

In Figs.~\ref{fig:QuaRoads} and \ref{fig:QuaRiverRoads}, we present the tubular structure centerline tracing results from  the Aniso, Progressive, FSR, Group-Angle  and the proposed Group-FSR models on road and river images.  In these experiments, the color images are first converted into gray levels in a preprocessing step, such that the cost function $\mathfrak{E}$ is extracted from these gray images. However, we still draw the tubular structure tracking results on the original color images for better visualization. 
In Fig.~\ref{fig:QuaRoads}, we show the qualitative results on aerial images of road networks~\cite{turetken2016reconstructing}. The main difficulty usually lies at the complicated background such as buildings which may also produce strong tubular appearance features. From columns $2$ to $5$ of Fig.~\ref{fig:QuaRoads}, one can observe the shortcuts occur when implementing the existing  models. While for the results from the proposed model, the objective structures can be correctly traced thanks to the proposed edge weights construction way. 
In Fig.~\ref{fig:QuaRiverRoads}, we demonstrate the extracted paths indicated by red lines on road and river patches from satellite images. The target tubular structures surrounded by complicated background also appear very long and have strong tortuosity. The tracing results derived from the existing results are shown in columns $2$ to $5$, from which we observe  paths with shortcuts. In contrast, the proposed Group-FSR model can follow the road and river structures successfully.

\begin{table*}[t]
\centering
\caption{The values of  $\mathcal J$  for evaluating the performance of the considered models on synthetic images shown in Fig.~\ref{fig:SyntheticImages}}
\label{tab:Synthetic}
\setlength{\tabcolsep}{8pt}
\renewcommand{\arraystretch}{2}
\begin{tabular}{c|| c c c c c c}
\shline
\multicolumn{1}{l}{Images}  &Aniso  &Progressive &FSR  & Group-Angle & Proposed Group-FSR &Proposed Group-FE\\ 
\hline   
\multirow{1}{*}{Column 1}  &$40.64\%$ &$41.41\%$   & $45.67\%$ &$42.64\%$ &$100.00\%$ &$100.00\%$\\
\multirow{1}{*}{Column 2} &$74.86\%$ &$66.86\%$ &$57.14\%$ &$75.10\%$  &$100.00\%$ &$100.00\%$\\ 
\multirow{1}{*}{Column 3} &$36.30\%$ &$16.53\%$   &$65.30\%$ &$68.09\%$  &$100.00\%$ &$100.00\%$\\
\multirow{1}{*}{Column 4}  &$52.71\%$ &$40.09\%$   &$52.82\%$ &$53.03\%$  &$100.00\%$ &$100.00\%$\\
\shline
\end{tabular}
\end{table*}

\begin{table*}[t]
\centering
\caption{Average values of $\cJ$  for evaluating the performance of the considered models in artery vessels tracing  on retinal images from the DRIVE and IOSTAR datasets}
\label{tab:AverageRetina}
\setlength{\tabcolsep}{8pt}
\renewcommand{\arraystretch}{2}
\begin{tabular}{c|c c c c c c}
\shline
\multicolumn{1}{l}{Datasets} &Aniso &Progressive   &FSR  & Group-Angle & Proposed Group-FSR & Proposed Group-FE\\ 
\hline
\multirow{1}{*}{DRIVE}  &$52.26\%$  &$54.92\%$   & $48.01\%$ &$84.01\%$ &$\mathbf{98.50\%}$ &$97.22\%$\\ 
\multirow{1}{*}{IOSTAR} &$67.09\%$ &$74.54\%$  &$78.71\%$ &$85.62\%$  &$\mathbf{98.43\%}$ &$96.79\%$ \\
\shline
\end{tabular}
\end{table*} 

\begin{figure*}[t]
\centering
\includegraphics[width=8.5cm]{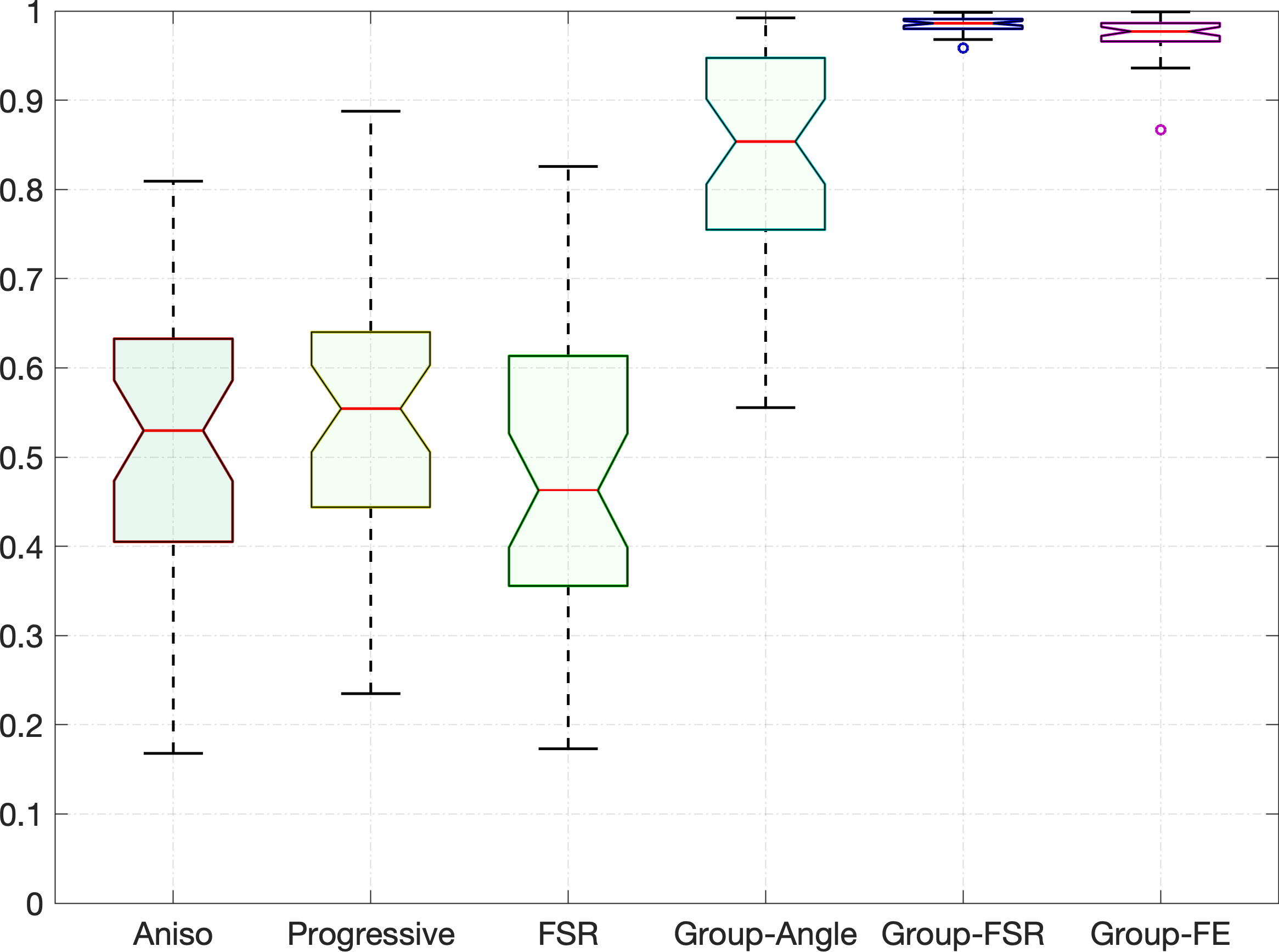}\quad\includegraphics[width=8.5cm]{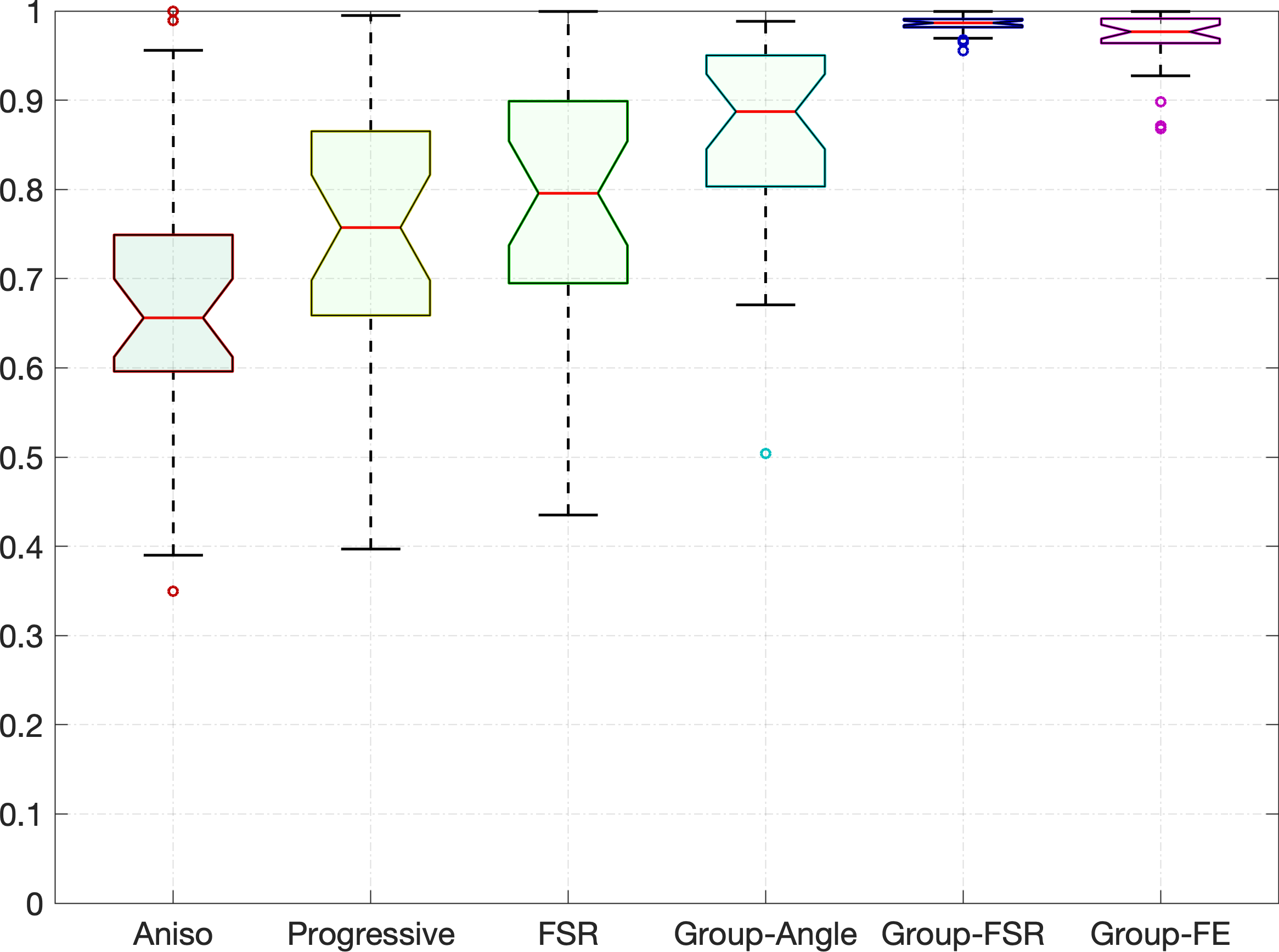}
\caption{Box plots of $\mathcal J$ for different methods on retinal images, where the left and right figures are  the  evaluation results corresponding to the DRIVE and IOSTAR datasets, respectively}
\label{fig:boxPlots}	
\end{figure*}

\subsection{Quantitative Evaluation Results}
\label{subsec:Quan}
The qualitative comparisons on synthetic images, retinal vessel images and natural images presented in Section~\ref{sec:QualitativeExp} prove that the proposed Group-FSR and Group-FE models indeed obtain promising results. In order to evaluate the proposed models in a rigorous and convincing manner, here we run the quantitative comparisons on the synthetic images and retinal images, respectively.
The accuracy that measures the performance of  the tested models is carried out by a score $\cJ$ defined as follows:
\begin{equation}
\label{eq:accuracyscore}
\cJ=\frac{\#|S\cap GT|}{\#|S|},
\end{equation}
where $S$ is the set of grid points passed through by the evaluated paths, $GT$ denotes the region of the ground truth, and $\#|S|$ stands for the elements involved in the set $S$. The accuracy score $\mathcal J$ is ranged within the interval [0,1], where higher values of $\cJ$ means better performance on tracing tubular structures.

We first present the performance of the compared models on the synthetic images shown in Fig.~\ref{fig:SyntheticImages}. The ground truth $GT$ is set as a binary segmentation of all pixels corresponding to the target structures. The accuracy scores $\cJ$ computed from the evaluated paths are shown in Table~\ref{tab:Synthetic}. One can point out that proposed Group-FSR and Group-FE models achieve the best performance among all the considered models. 

In Table~\ref{tab:AverageRetina}, we show the quantitative evaluation results on the retinal images from the datasets of DRIVE~\cite{staal2004ridge} and IOSTAR~\cite{zhang2016robust}. Tracing artery vessels from retinal images providing that only few user-provided points are given is a challenging task, thus can well measure the performance of the considered models. For a quantitative evaluation, we compare our proposed Group-FSR and Group-FE models to the Aniso, Progressive, FSR and Group-Angle models, in the sense of the accuracy score $\cJ$ defined in Eq.~\eqref{eq:accuracyscore}. We have extracted the ground truth regions for each  tested artery vessel from the artery-vein ground truth images. Note that each artery-vein ground truth image classifies each grid point belonging to either artery, vein or background. We made use of $394$ artery vessels sampled from two datasets, where most of the major artery vessels in each retinal image are taken into account for evaluation.  Among all the tests, we provide only $2$ points for $321$ vessels and the remaining cases require $3$ points. 
Again, we note that each individual artery vessel is tracked using the entire image. In each test, all the evaluated models are under the same user-supplied points. In Table.~\ref{tab:AverageRetina}, we illustrate the average accuracy scores for different models.  In this table, we observe that the proposed Group-FSR and Group-FE models demonstrate large gaps to the other tested models, proving the advantages of the proposed models. In addition, we can see that in each dataset the average values of  $\cJ$ for the proposed Group-FSR and Group-FE models are similar to each other and achieve around $12\%$ higher than the Group-Angle model. Note that we utilized the same node and edge sets to construct the graphs for the Group-Angle model and the proposed models, except for the computation of the edge weights. 

Finally,  we show more statistical results on the accuracy scores $\cJ$ for all the tested models through the tool of box plots, as depicted in Fig.~\ref{fig:boxPlots}. The left and right columns in Fig.~\ref{fig:boxPlots} correspond to the results on the DRIVE and IOSTAR datasets, respectively.  The results shown in this figure again  prove that the proposed models outperform the compared state-of-the-art models in both robustness and accuracy.


\section{Conclusions}
\label{sec:conclusion}
In this paper, we propose a new minimal paths-based approach for the delineation of tubular structure centerlines by grouping  a  set of prescribed trajectories based on the Dijkstra's shortest path method. A crucial point for the proposed method concentrates on the use of data-driven curvature-penalized geodesic paths used to fit the lost vessel segments between prescribed trajectories. Accordingly, the proposed model is able to blend the benefits from the prescribed tubular trajectories, the curvature-penalized geodesic paths and the global optimality of Dijkstra's algorithm.  The experimental results prove that the proposed models (involving both the Group-FSR or Group-FE models)  indeed outperform state-of-the-art minimal path-based tubular structure tracking models such as the anisotropic tubular geodesic model~\cite{benmansour2011tubular}, the progressive minimal path model~\cite{liao2018progressive}, the FSR model with curvature regularization~\cite{duits2018optimal}. The future work can be devoted to developing  algorithms for   automatic extraction of tubular tree structures based on the proposed Group-FSR or Group-FE models.


%



\section*{Acknowledgment}
The authors would like to thank all the anonymous reviewers for their invaluable  suggestions to improve this manuscript. This work is in part supported by the National Natural Science Foundation of China (NOs.62102210, 61902224, 62171125),  the Shandong Provincial Natural Science Foundation (NO.ZR202102240225), the Shanghai Sailing Program (NO.20YF1401500), the Fundamental Research Funds for the Central Universities (NO.2232020D-35), the Basic Research Enhancement Program (NO.2021JC04010), the French government under management of Agence Nationale de la Recherche as part of the ``Investissements d'avenir'' program, reference ANR-19-P3IA-0001 (PRAIRIE 3IA Institute) and by the Young Taishan Scholars (NO.tsqn201909137).
\bibliographystyle{IEEEbib}
\bibliography{pGTubular}

\ifCLASSOPTIONcaptionsoff
  \newpage
\fi



%

%

%
\begin{IEEEbiographynophoto}{Li Liu} received her Ph.D. degree in computer science and technology from Southeast University, Nanjing, China, in 2019. From 2017 to 2019, she was a joint training PhD student in Paris Dauphine University, PSL Research University.  Now she is working in Shandong Artificial Intelligence Institute, Qilu University of Technology (Shandong Academy of Sciences). Her research interests include geodesic model and its applications in image analysis and medical imaging.
\end{IEEEbiographynophoto}

\begin{IEEEbiographynophoto}{Da Chen}received his Ph.D degree supervised by Prof. Laurent D. Cohen in applied mathematics from CEREMADE, University Paris Dauphine, PSL Research University, Paris, France, in 2017. From 2017 to 2019, he worked as a post-doctoral researcher at CEREMADE, University Paris Dauphine, and also at Centre Hospitalier National d’Ophtalmologie des Quinze-Vingts, Paris, France.  Now he is working in Shandong Artificial Intelligence Institute, Qilu University of Technology (Shandong Academy of Sciences). His research interests include variational methods, partial differential equations, machine learning, and geometric methods as well as their applications in computer vision and image analysis, such as minimal geodesic paths, active contours, image/volume segmentation, tubular structure tracking, surface reconstruction and medical image registration. 
\end{IEEEbiographynophoto}

\begin{IEEEbiographynophoto}{Minglei Shu}
received his B.S. degree in automation from Shandong University in 2003, the M. Sc. degree in power electronics and power transmission from Shandong University in 2006 and Ph.D. degree in communication and information systems from Shandong University in 2016. Currently, he is working at Shandong Artificial Intelligence Institute, Qilu University of Technology (Shandong Academy of Sciences), and Deputy Director of Shandong Artificial Intelligence Institute. His research interests mainly include computer vision, medical image segmentation, 3D blood vessel segmentation and tracking, IoT medical care as well as wireless sensor networks. 		
\end{IEEEbiographynophoto}

\begin{IEEEbiographynophoto}{Baosheng Li} is a distinguished professor at the Cancer Hospital of Shandong First Medical University. He received his Ph.D. degree in biomedical engineering from Southeast University in 2004 and worked as a visiting scholar at the Medical School of Maryland University from 1999 to 2000. His research focuses on the precision radiotherapy, radioimmunotherapy, and application of organ motion analysis and functional imaging in radiotherapy of cancers.
\end{IEEEbiographynophoto}

\begin{IEEEbiographynophoto}{Huazhong Shu} received the B.S. degree in Applied Mathematics from Wuhan University, China, in 1987, and a Ph.D. degree in Numerical Analysis from the University of Rennes $1$, Rennes, France in 1992. He is a professor of the LIST Laboratory and the Codirector of the CRIBs. His recent work concentrates on the image analysis, pattern recognition and fast algorithms of digital signal processing. Dr. Shu is a senior member of IEEE Society.
\end{IEEEbiographynophoto}

\begin{IEEEbiographynophoto}{Michel Paques}
received the MD in ophthalmology and Ph.D. degrees in biomedical research from University of Paris 7, France, in 1994 and 2000, respectively. He joined the Vision Institute in 2002 and then the Quinze-Vingts hospital in Sorbonne Université in 2008 as a Professor of ophthalmology. He co-founded and heads the Paris group lab (http://parisgroup.webstarts.com/index.html) dedicated to developing novel approaches for retinal imaging in patients. He pioneered the modelization of ocular circulation and contributed to the understanding of neuronal damage during eye diseases such as photoreceptor misalignment and acute ischemia.	
\end{IEEEbiographynophoto}

\begin{IEEEbiographynophoto}{Laurent D. Cohen} was student at the Ecole Normale Superieure, rue d'Ulm in Paris, France, from 1981 to 1985. He received the Master's and Ph.D. degrees in applied mathematics from University of Paris 6, France, in 1983 and 1986, respectively. He got the Habilitation \'a diriger des Recherches from University Paris $9$ Dauphine in 1995. From 1985 to 1987, he was member at the computer graphics and image processing group at Schlumberger Palo Alto Research, Palo Alto, California, and Schlumberger Montrouge Research, Montrouge, France, and remained consultant with Schlumberger afterward. He began working with INRIA, France, in 1988, mainly with the medical image understanding group EPIDAURE. He obtained in 1990 a position of Research Scholar (Charge then Directeur de Recherche 1st class) with the French National Center for Scientific Research (CNRS) in the Applied Mathematics and Image Processing group at CEREMADE, Universite Paris Dauphine, Paris, France. His research interests and teaching at university are applications of partial differential equations and variational methods to image processing and computer vision, such as deformable models, minimal paths, geodesic curves, surface reconstruction, image segmentation, registration and restoration. He is currently or has been editorial member of the Journal of Mathematical Imaging and Vision, Medical Image Analysis and Machine Vision and Applications. He was also member of the program committee for about 50 international conferences. He has authored about 260 publications in international Journals and conferences or book chapters, and has 6 patents. In 2002, he got the CS 2002 prize for Signal and Image Processing. In 2006, he got the Taylor \& Francis Prize: ``2006 prize for Outstanding innovation in computer methods in biomechanics and biomedical engineering." He was 2009 laureate of Grand Prix EADS de l’Academie des Sciences in France. He was promoted as IEEE Fellow 2010 for contributions to computer vision technology for medical imaging.
\end{IEEEbiographynophoto}


\end{document}